%% file: main.tex
\documentclass{article} 
\usepackage{iclr2027_conference,times}

\input{math_commands.tex}

\usepackage{hyperref}
\usepackage{url}

\usepackage[utf8]{inputenc} 
\usepackage{booktabs}       
\usepackage{amsfonts}       
\usepackage{xcolor}         
\usepackage{graphicx}
\usepackage{multirow}
\usepackage{comment}
\usepackage{enumitem}
\usepackage[most]{tcolorbox}
\usepackage[table]{xcolor}
\usepackage{array}

\title{APT: Feedback-Driven Agentic Post-Training for Text-to-Image Generation}

\author{%
   Sirui Zhang$^{1,2}$\thanks{Work done during internship in MSRA. \hspace{1.5mm}$^\dagger$~Corresponding author.}\hspace{1.5mm}\hspace{3mm} Fangyun Wei$^{2}$\hspace{3mm} Dong Chen$^{2}$\hspace{3mm} Jinjing Zhao$^{3}$\hspace{3mm} Yitong Wang$^{4}$ \\
  \textbf{Yunuo Chen$^{5}$\hspace{3mm} Baining Guo$^{2}$\hspace{3mm} Zheng-Jun Zha$^{1\dagger}$} \\
  {$^{1}$~University of Science and Technology of China}  \qquad {$^{2}$~Microsoft Research Asia}  \vspace{1.5pt} \\ 
  {$^{3}$~University of Sydney} \qquad
  {$^{4}$~Fudan University} \qquad
  {$^{5}$~Shanghai Jiao Tong University} \\
  \texttt{\{zsr200901, zhazj\}@mail.ustc.edu.cn} \\
  \texttt{\{fawe, doch, bainguo\}@microsoft.com} \\
}

\arxivcopy
\begin{document}

\maketitle

\input{sections/0-abstract}
\input{sections/1-introduction}
\input{sections/3-method}
\input{sections/4-experiment}

\input{sections/5-conclusion}

\newpage

\bibliography{iclr2027_conference}
\bibliographystyle{iclr2027_conference}

\input{sections/X-appendix}

\end{document}

%% file: math_commands.tex
\usepackage{amsmath,amsfonts,bm}

\def\eqref#1{equation~\ref{#1}}

\def\1{\bm{1}}

\DeclareMathAlphabet{\mathsfit}{\encodingdefault}{\sfdefault}{m}{sl}
\SetMathAlphabet{\mathsfit}{bold}{\encodingdefault}{\sfdefault}{bx}{n}



%% file: sections/0-abstract.tex
\begin{figure}[!h]
    \centering
    \vspace{-6mm}
    \includegraphics[width=0.99\textwidth]{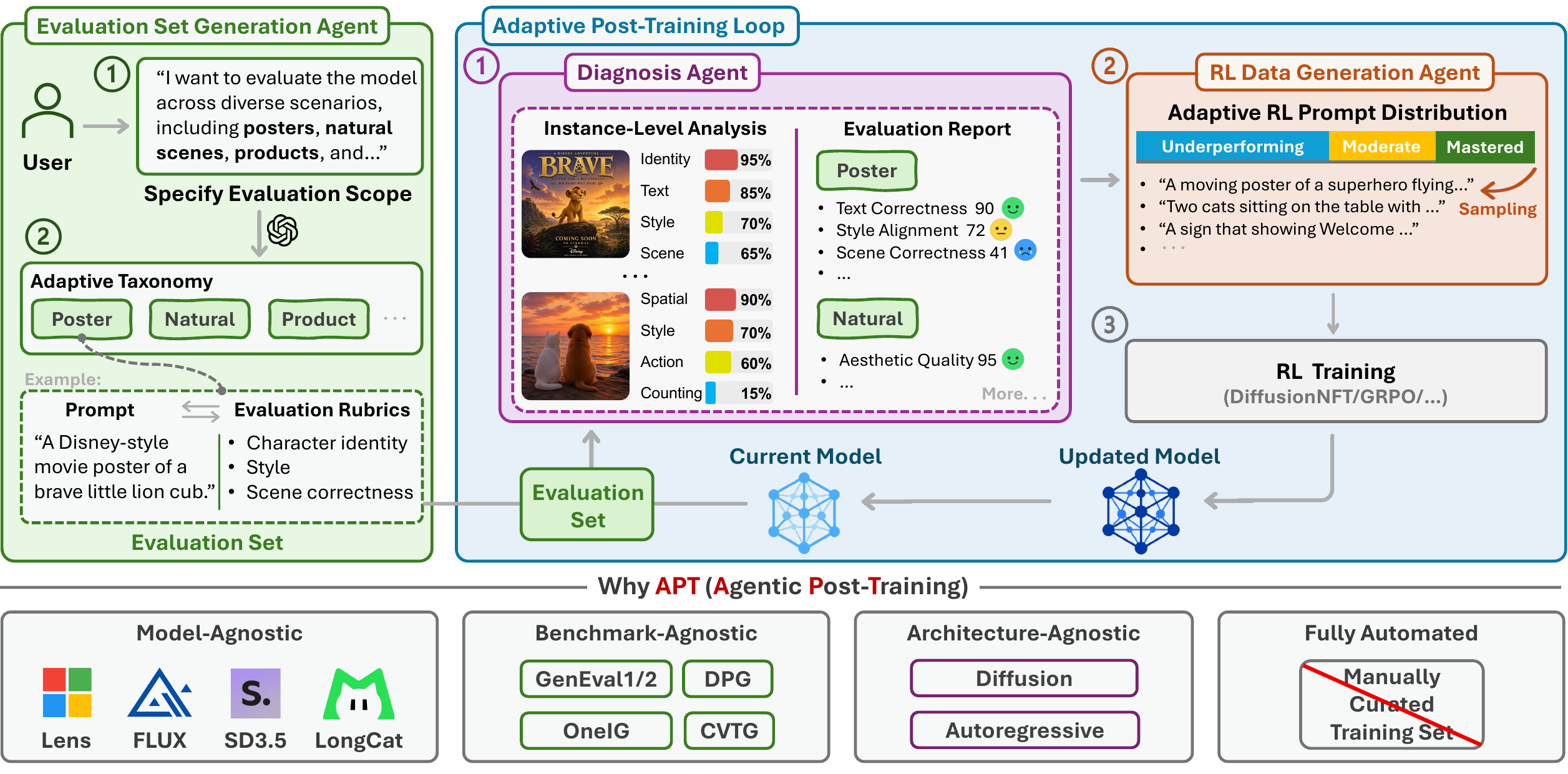}
    \vspace{-3mm}
    \caption{\textbf{APT (Agentic Post-Training)} turns post-training into a fully automated, evaluation-driven adaptive loop.
Given a user-specified evaluation scope, the \textcolor[HTML]{3A7D23}{\textbf{Evaluation Set Generation Agent}} constructs a diverse evaluation set consisting of \emph{prompt--evaluation rubric pairs}; the \textcolor[HTML]{78206E}{\textbf{Diagnosis Agent}} evaluates the current model on these pairs to identify its strengths and failure modes; and the \textcolor[HTML]{C04F15}{\textbf{RL Data Generation Agent}} adaptively samples next-round training prompts, prioritizing underperforming capabilities.
The updated model is then re-evaluated in the next round, allowing the training distribution to continuously adapt to the model's evolving capability profile.
This design eliminates the need for manually curated post-training data and makes APT \textit{model-agnostic}, \textit{benchmark-agnostic}, \textit{architecture-agnostic}, and \textit{fully automated}.}
    \vspace{-2mm}
    \label{fig:teaser}
\end{figure}

\begin{abstract}
Post-training is crucial for improving text-to-image models, yet existing pipelines typically rely on manually curated training data and fixed training distributions.
We introduce APT (Agentic Post-Training), a fully automated framework that adaptively post-trains a base model according to its evolving capability profile.
APT coordinates three agents to construct evaluation sets from user requirements, diagnose model-specific strengths and failure modes, and generate reinforcement learning data with an adaptive prompt distribution that emphasizes underperforming capabilities.
By repeatedly evaluating, diagnosing, and updating the generator, APT lets the current model's behavior shape its next training tasks, forming a feedback loop inspired by recursive self-improvement.
For general-purpose settings without user-specified evaluation scenarios, we further develop an evaluation harness that enables broad and diverse evaluation set generation.
Experiments across multiple base models, benchmarks, and generation architectures demonstrate that APT is model-agnostic, benchmark-agnostic, architecture-agnostic, and fully automated.
\end{abstract}

%% file: sections/1-introduction.tex
\section{Introduction}

Recent text-to-image (T2I) models have achieved remarkable progress in visual quality, prompt following, text rendering, compositional generation, and controllability~\citep{t2i-flux-2,t2i-gpt-image-2,t2i-nanobanana-2,t2i-seedream-5-pro,t2i-qwen-image-2}.
Beyond large-scale pre-training, post-training has become increasingly important for correcting specific failure modes and aligning model behavior with desired objectives~\citep{rl-qwen-image-2,t2i-lens,rl-z-image,t2i-sefi}.
Existing post-training pipelines, however, typically rely on manually constructed training distributions: practitioners identify model weaknesses, collect or synthesize corresponding training data, and repeatedly adjust the data distribution based on empirical observations~\citep{rl-qwen-image-2,t2i-lens,t2i-sefi,t2i-x-omni}.
Such a workflow is labor-intensive and difficult to scale as models, benchmarks, and user requirements continue to diversify.

More fundamentally, we argue that post-training should be \emph{model-adaptive}.
Different base models exhibit substantially different capability distributions: a model that performs well on text rendering may struggle with counting or spatial relations, while another may exhibit the opposite behavior.
Consequently, a post-training dataset that is effective for one model is not necessarily optimal for another.
We can view each base model $M$ as inducing its own capability and failure distribution $p_M$ over generation tasks.
Effective post-training should therefore identify where the current model underperforms, sample training examples accordingly~\citep{chen2026unified}, and continuously adapt the sampling distribution as the model improves~\citep{li2026curriculum}.
In other words, the key question is not only \emph{how to optimize a model}, but also \emph{what the current model should be trained on}. This perspective naturally connects post-training to recursive self-improvement (RSI), in which the evaluation results of the updated model are used to determine the training tasks for the next round~\citep{zelikman2023self,zweiger2026self,huang2026r}.

To address this problem, we introduce \textbf{APT (Agentic Post-Training)}, a fully automated framework that turns post-training into an adaptive, evaluation-driven self-improvement loop, as illustrated in Figure~\ref{fig:teaser}.
Given a T2I base model and an evaluation requirement, APT automatically constructs an evaluation set of prompt--rubric pairs, diagnoses the current model's strengths and weaknesses, and generates model-specific RL training data.
The updated model is then re-evaluated, allowing its evolving behavior to continuously reshape the training distribution for the next round.
Rather than relying on a fixed, manually curated dataset, APT lets the current model determine what should be learned next.

As shown in Figure~\ref{fig:teaser}, APT is built around three collaborative agents.
The \textit{Evaluation Set Generation Agent} converts a high-level evaluation scope into a structured evaluation suite with an adaptive taxonomy, prompts, and task-specific rubrics.
The \textit{Diagnosis Agent} evaluates the current model at the instance and category levels and summarizes systematic strengths and failure modes.
The \textit{RL Data Generation Agent} then converts this diagnosis into an adaptive prompt distribution that prioritizes underperforming capabilities while retaining moderate and mastered ones, producing model-specific data for RL-based post-training~\citep{diffusionnft,flow-grpo,grpo}.
Repeated evaluation, diagnosis, and adaptive sampling provide an implicit mechanism for discovering and tracking the evolving model-dependent distribution $p_M$ without explicitly estimating it.

APT also supports general-purpose post-training when no evaluation scope is specified.
For this setting, we develop an \textbf{evaluation harness} that guides the Evaluation Set Generation Agent to systematically construct a broad and diverse evaluation set spanning common generation scenarios and capability dimensions~\citep{rubricrl}.

The resulting framework is \textit{model-agnostic}, \textit{benchmark-agnostic}, \textit{architecture-agnostic}, and \textit{fully automated}, and we validate these properties across multiple T2I models, generation architectures, and benchmarks~\citep{t2i-lens,t2i-sd3.5-m,t2i-flux1.dev,t2i-longcat-image,t2i-janus-pro}.
APT consistently improves diverse base models and outperforms post-training with a carefully handcrafted RL training set.
Moreover, transferring APT-generated training data across models is consistently inferior to adaptively generating data for the target model, supporting our central hypothesis that effective post-training distributions are model-specific and should evolve together with the model.
Together, these results establish agentic, evaluation-driven adaptation as a practical step toward recursive self-improvement for text-to-image generation.

%% file: sections/3-method.tex
\section{Method}

\textbf{Overview.}
Figure~\ref{fig:teaser} provides an overview of APT.
The Evaluation Set Generation Agent is invoked once before post-training to construct the fixed evaluation set $\mathcal{E}$ of prompt--rubric pairs (Section~\ref{sec:eval-set-agent}).
Users may specify an evaluation scope; when no scope is provided, we use our general-purpose evaluation harness to ensure broad and comprehensive coverage.
APT then enters an iterative post-training loop.
At each round, the Diagnosis Agent (Section~\ref{sec:diagnosis-agent}) evaluates the current model and produces an evaluation report, which is passed to the RL Data Generation Agent (Section~\ref{sec:data-agent}) to generate training data for the next RL update.
This RSI-inspired loop repeats after each update, enabling the training distribution to adapt to the model's evolving weaknesses.
As described in Section~\ref{sec:training-overall}, APT is model-agnostic, benchmark-agnostic, and architecture-agnostic, and can therefore be applied to diverse text-to-image models and evaluation settings.

\textbf{Round-Wise Learning Objective.}
At the beginning of round $t$, $M_t$ denotes the current text-to-image generator; $p$ and $x$ denote a text prompt and a generated image, respectively.
Let $D_t$ be the generated dataset of training prompts for this round, and write $(p,R)\sim D_t$ to denote sampling a prompt--rubric-set pair from this dataset.
For a prompt $p_i$, let $R_i$ be its associated rubric set and $\rho\in R_i$ an individual evaluation criterion.
Both diagnosis and training use GPT-4.1-mini as vision-language evaluator, denoted by $V$.
Its binary output $V(x,p,\rho)\in\{0,1\}$ is $1$ if image $x$ satisfies criterion $\rho$ for prompt $p$, and $0$ otherwise.
For an image $x_i$ generated from $p_i$, the training reward is the arithmetic mean of these rubric-level scores:
\[
r(x_i,p_i)=\frac{1}{|R_i|}\sum_{\rho\in R_i}V(x_i,p_i,\rho).
\]
Each prompt's rubric set is fixed and shared across all its generated images within a rollout round.
The evaluator's scoring instructions are provided in Appendix~\ref{sec:prompt-reward}.
The round-wise expected-reward objective is
\begin{equation}
J(M_t)=\mathbb{E}_{(p,R)\sim D_t,\,x\sim M_t(\cdot\mid p)}\!\left[r(x,p)\right].
\label{eq:round-objective}
\end{equation}
Starting from $M_t$, the configured RL algorithm takes a finite number of optimization steps aimed at increasing $J(M_t)$, with $D_t$ and the reward procedure fixed within the round.

\textbf{Recursive Data--Model Coupling.}
APT couples an \emph{outer loop of diagnostic-driven task adaptation} with an \emph{inner loop of RL-based model optimization}.
At round $t$, $\mathcal{B}_t\subset\mathcal{E}$ is a sampled diagnostic batch of prompt--rubric pairs, and $\mathcal{R}_t$ denotes the resulting rubric-level evaluation report for $M_t$.
Let $\operatorname{Diag}$ and $\mathcal{G}$ denote the diagnosis and data-generation procedures, respectively, and let $\mathcal{U}$ denote the RL update operator.
The coupled update is
\begin{equation}
\begin{aligned}
\mathcal{R}_t &= \operatorname{Diag}(M_t,\mathcal{B}_t),
\qquad D_t=\mathcal{G}(\mathcal{R}_t), \\
M_{t+1} &= \mathcal{U}(M_t,D_t;V).
\end{aligned}
\label{eq:recursive-post-training}
\end{equation}
The update operator $\mathcal{U}$ returns $M_{t+1}$ by applying the configured finite number of RL steps to $M_t$, with the aim of maximizing $J(M_t)$ in Equation~\ref{eq:round-objective}.
Re-evaluating $M_{t+1}$ yields a diagnostic report $\mathcal{R}_{t+1}$ that guides the generation of training prompts $D_{t+1}$ for the next RL update.
We next describe the three agents that instantiate this feedback-driven post-training loop.

\begin{figure}[!t]
    \centering
    \includegraphics[width=0.99\textwidth]{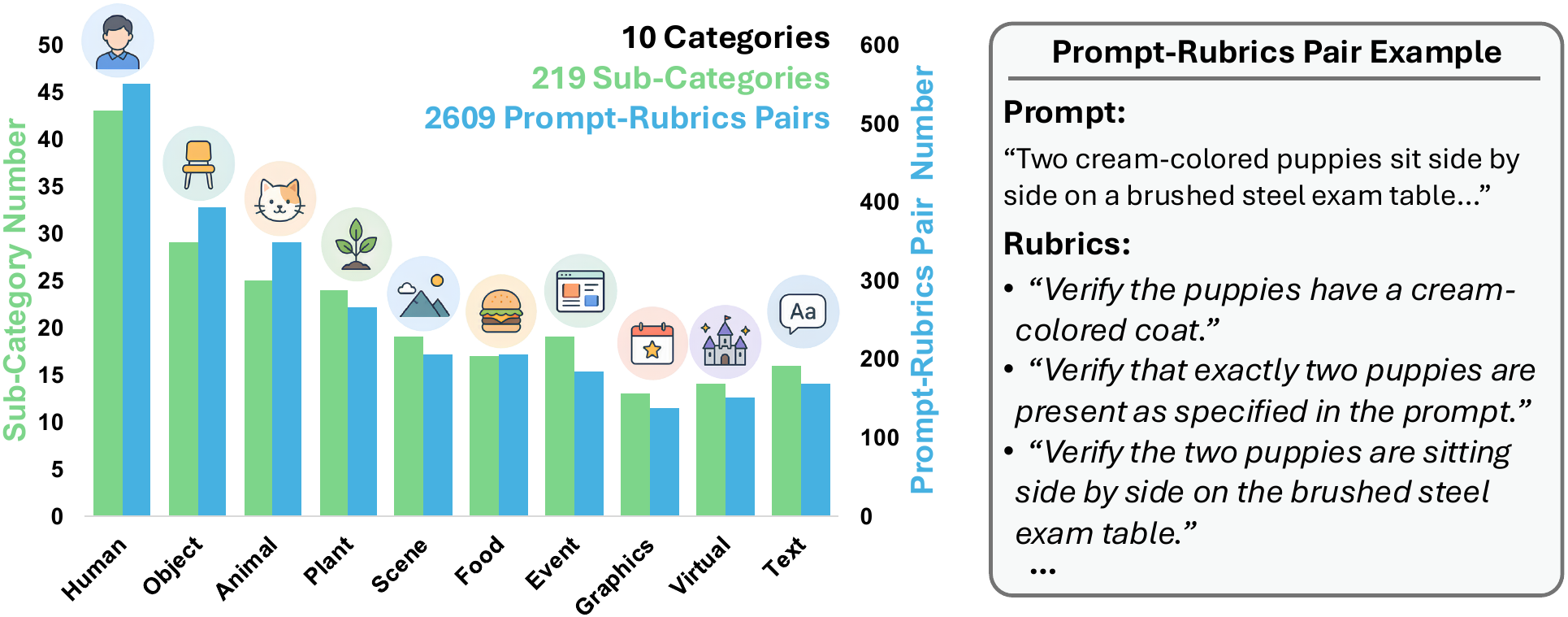}
    \vspace{-3mm}
    \caption{\textit{Left}: Statistics of the sub-categories within each category and the prompt-rubrics pairs within each category. All prompt-rubrics pairs are automatically generated by our Evaluation Set Generation Agent, providing comprehensive coverage for evaluating the current T2I model. \textit{Right}: An example prompt-rubrics pair.}
    \vspace{-6mm}
    \label{fig:agent-1}
\end{figure}

\subsection{Evaluation Set Generation Agent}
\label{sec:eval-set-agent}
The Evaluation Set Generation Agent is responsible for constructing the evaluation data used to diagnose the current model before the post-training stage.
When a user specifies an evaluation scope, the agent directly generates evaluation examples tailored to the requested scenarios.
For the more general setting where no evaluation scope is provided, we further design a \textit{general-purpose evaluation harness} that guides the agent to systematically construct a broad-coverage evaluation set spanning diverse image-generation scenarios and capabilities.
Each evaluation example consists of an image-generation prompt paired with its corresponding prompt-specific evaluation rubrics.
The evaluation set is constructed through a multi-step generation pipeline consisting of scenario generation, prompt generation, and rubric generation.

\textbf{Step 1: Scenario Generation.}
We begin with a broad set of common image-generation scenario categories, including Human, Object, Animal, Plant, Scene, Food, Event, Fictional World, Text, and UI and Graphic Design.
For each category, we prompt an LLM to generate dozens of fine-grained sub-categories and corresponding items.
For example, the \textit{Human} category may include sub-categories such as \textit{Race}, \textit{Occupation}, and \textit{Gender}, while the \textit{Occupation} sub-category may further contain items such as \textit{Researcher} and \textit{Doctor}.
This process produces a hierarchical scenario taxonomy that provides broad and structured coverage of diverse image-generation scenarios.

\textbf{Step 2: Prompt Generation.}
Given the items derived from the scenario taxonomy, we further enrich them with a set of descriptive dimensions to generate diverse image-generation prompts.
These descriptions capture common compositional and semantic requirements, such as attributes, spatial relationships, object counts, interactions, and colors.
For each item, we randomly sample one to four descriptions and prompt GPT-4.1, together with a predefined system prompt described in Appendix~\ref{sec:prompt-eval-prompt}, to synthesize the final image-generation prompt.
For example, an item such as \textit{Doctor} can be combined with descriptions of color, interaction, and spatial relationship to produce a more compositional and challenging prompt.
This design increases both the diversity and coverage of the generated evaluation prompts.

\textbf{Step 3: Rubric Generation.}
For each generated prompt, we further construct prompt-specific evaluation rubrics~\citep{rubricrl}.
Concretely, we feed the prompt, together with the rubric-generation system prompt described in Appendix~\ref{sec:prompt-rubric}, into GPT-4.1-mini to generate up to 10 prompt-specific rubrics.
\textit{These rubrics specify the aspects that should be checked for the given prompt}, such as identity consistency, text correctness, style alignment, scene correctness, spatial relations, or counting accuracy, depending on the prompt content.
As a result, each evaluation example is represented as a prompt paired with its corresponding rubrics, enabling fine-grained and task-aware diagnosis of model behavior.

Overall, the Evaluation Set Generation Agent transforms a high-level evaluation requirement into a diverse evaluation set containing 2,609 prompt--rubrics pairs. Detailed statistics are provided in Figure~\ref{fig:agent-1}. This evaluation set serves as the foundation for the subsequent Diagnosis Agent, which analyzes model performance and drives the adaptive post-training loop.

\subsection{Diagnosis Agent}
\label{sec:diagnosis-agent}

The Diagnosis Agent analyzes the performance of the current model on the evaluation set and summarizes its strengths and weaknesses into a structured evaluation report.
At each post-training round $t$, we randomly sample a batch $\mathcal{B}_t=\{(p_i,R_i)\}_{i=1}^{N}\subset\mathcal{E}$ of prompt--rubric pairs, where $N$ is the diagnostic batch size ($120$ by default).
The diagnosis process consists of two stages: \textit{instance-level analysis} and \textit{evaluation report construction}.

\textbf{Instance-Level Analysis.}
Within round $t$, let $K_i=|R_i|$ be the number of rubrics for the $i$-th sampled prompt $p_i$, and let $\rho_{ij}$ denote its $j$-th rubric.
We generate an image $x_i^{(t)}$ with the current model $M_t$ and use the same evaluator $V$ as in the training reward to obtain the rubric score $s_{ij}^{(t)}$:
\begin{equation}
x_i^{(t)}\sim M_t(\cdot\mid p_i),
\qquad s_{ij}^{(t)}=V(x_i^{(t)},p_i,\rho_{ij})\in\{0,1\}.
\label{eq:rubric-feedback}
\end{equation}
This yields a set of prompt-specific rubric-level evaluation results $E_i^{(t)}=\{(\rho_{ij},s_{ij}^{(t)})\}_{j=1}^{K_i}$, providing a fine-grained diagnosis of the current model for prompt $p_i$.

\textbf{Evaluation Report Construction.}
After processing all $N$ sampled prompt--rubric pairs, the Diagnosis Agent constructs the report $\mathcal{R}_t=\{(p_i,E_i^{(t)})\}_{i=1}^{N}$, instantiating the diagnosis procedure $\operatorname{Diag}$ in Equation~\ref{eq:recursive-post-training}.
Each entry in $\mathcal{R}_t$ contains the original prompt together with its rubric-level evaluation results, where each result consists of an evaluation rubric and its corresponding binary score.
The resulting report captures the current model's successes and failures across diverse generation capabilities and is subsequently passed to the RL Data Generation Agent to guide the construction of training data for the subsequent RL update.

\subsection{RL Data Generation Agent}
\label{sec:data-agent}

The RL Data Generation Agent takes the evaluation report produced by the Diagnosis Agent as input and generates training prompts for the subsequent RL update.
Its main role is to identify recurring failure patterns in the report, summarize the capabilities that should be improved, and translate these findings into a model-specific training distribution.

We use the evaluation report as an empirical characterization of the current model's capability distribution.
This is feasible for several reasons.
First, the evaluation set is designed to provide broad coverage over diverse generation scenarios and capabilities.
As a result, the sampled evaluation examples probe many different regions of the model's capability space rather than concentrating on a narrow benchmark-specific domain.
Second, each prompt is associated with multiple prompt-specific rubrics, which provide fine-grained feedback on \emph{what} the model succeeds or fails on.
Therefore, the report contains not only whether a generation is successful, but also structured evidence about the underlying capability responsible for the success or failure.
Third, modern language models such as GPT-5.4 have strong summarization and reasoning capabilities, allowing them to compare observations across many evaluation examples and identify recurring patterns that may not be obvious from any individual sample.
For instance, prompts from different semantic categories may repeatedly fail for the same underlying reason, such as accurate counting, rendering text on objects, or preserving specific attributes.
By aggregating these cross-example patterns, the agent can infer which capabilities are consistently underperforming and which are already handled reliably.
Finally, because the evaluation report is regenerated from the current model at every post-training round, the inferred capability profile naturally changes as the model improves.
In this sense, the Diagnosis Agent and RL Data Generation Agent together provide an implicit and continually updated estimate of the model's current performance distribution, without requiring an additional explicitly trained estimator.

Based on the summarized failure patterns, the RL Data Generation Agent generates $N_{\text{train}}$ new prompts for the current round's RL update, with $p_{t,k}^{\mathrm{train}}$ denoting the $k$-th generated prompt and $R_{t,k}^{\mathrm{train}}$ denoting its corresponding rubric set generated by GPT-4.1-mini using the same procedure and system prompt as for the evaluation set :
\begin{equation}
D_t=\mathcal{G}(\mathcal{R}_t)=\{(p_{t,k}^{\mathrm{train}}, R_{t,k}^{\mathrm{train}})\}_{k=1}^{N_{\text{train}}}.
\label{eq:adaptive-prompts}
\end{equation}
This instantiates the data-generation procedure $\mathcal{G}$ in Equation~\ref{eq:recursive-post-training}, conditioned on the current model's diagnostic report $\mathcal{R}_t$.
A clear majority of these prompts target recurring weak capabilities, while a smaller portion covers moderately performing and already-strong capabilities to preserve diversity and avoid over-concentrating on a single failure mode.
Since different base models exhibit different patterns in the same broad evaluation space, this process naturally produces model-specific training data.
Moreover, as the model evolves over successive rounds, the generated training distribution is updated accordingly.

We implement the RL Data Generation Agent with GPT-5.4.
The detailed instruction prompt used by the agent is provided in Appendix~\ref{sec:prompt-agent}.

\subsection{General, Adaptive, and Fully Automated Post-Training}
\label{sec:training-overall}

APT is designed as a general post-training framework that can be applied across different evaluation settings, base models, and generation architectures without task-specific or model-specific redesign. Specifically, APT is (1) \textit{benchmark-agnostic}: it does not require independently designed prompt sets for individual benchmarks, and a single APT run can be evaluated across multiple T2I benchmarks; (2) \textit{model-agnostic}: the same agentic workflow can be directly applied to diverse base models, including Lens~\citep{t2i-lens}, SD3.5~\citep{t2i-sd3.5-m}, FLUX~\citep{t2i-flux1.dev}, and LongCat-Image~\citep{t2i-longcat-image}; (3) \textit{architecture-agnostic}: APT is not tied to a particular image-generation formulation and is effective for both diffusion-based and autoregressive image-generation models~\citep{t2i-janus-pro,t2i-llamagen}; and (4) \textit{fully automated}: the evaluation results of the current model directly drive the generation of prompts for subsequent RL updates, eliminating the need for manually inspecting failures, designing training distributions, or curating model-specific post-training data~\citep{rl-qwen-image-2,t2i-x-omni}.

\textbf{Recursive Self-Improvement for Text-to-Image Generation.}
APT brings recursive self-improvement (RSI)~\citep{zelikman2023self,zweiger2026self} to text-to-image generation by making the generator's evolving capabilities drive the construction of its own training curriculum.
As formalized in Equation~\ref{eq:recursive-post-training}, the current model produces behavioral evidence that agents translate into targeted training tasks; RL updates the generator on these tasks, and the updated generator provides fresh evidence for the next round.
This creates a recursive coupling between model optimization and task construction: learning changes the model's capability profile, which in turn changes what the system trains on next.
Through this feedback loop, APT continually revises its training priorities in response to the generator's remaining and newly exposed weaknesses.
By automating diagnosis, curriculum construction, and RL updates, APT turns the generator's own successes and failures into actionable signals for directing subsequent improvement.
APT thus operationalizes RSI for T2I post-training through an adaptive curriculum that evolves alongside the model, replacing manual intervention in deciding what the model should learn next.

%% file: sections/4-experiment.tex
\section{Experiments}
\label{sec:exp}


\subsection{Experimental Setup}
\label{sec:exp-setup}

\begin{table*}[!t]
\vspace{-2mm}
\centering
\caption{Comparison of APT with the original base models and a handcrafted RL post-training baseline across four T2I models. The handcrafted RL set is manually curated by human experts to cover diverse generation scenarios, while APT uses the same model-agnostic agentic workflow for all models. APT consistently achieves the best average performance and improves multiple benchmarks with a single post-training run, demonstrating both stronger post-training effectiveness and benchmark-agnostic generalization. Detailed results for each benchmark are shown in Tables~\ref{tab:main_geneval}--\ref{tab:main_cvtg2k} in the Appendix~\ref{sec:benchmark-details}.}
\vspace{1mm}
\label{tab:main_all}
\small
\renewcommand{\arraystretch}{1.05}
\resizebox{\textwidth}{!}{%
\begin{tabular}{lcccccc}
\toprule
\multirow{2}{*}[-1pt]{\textbf{Model}} & \textbf{GenEval} & \textbf{OneIG-EN} & \textbf{GenEval2} & \textbf{DPG} & \textbf{CVTG-2K} & \multirow{2}{*}[-1pt]{\textbf{Average}} \\
\cmidrule(lr){2-6}
& Overall & Average & Soft-TIFA$_{\mathrm{GM}}$ & Overall & Word Accuracy & \\
\midrule
\multicolumn{7}{l}{\emph{\textbf{Lens}}} \\
Base Model & 0.702 & 0.521 & 29.06 & 85.93 & 0.480 & 57.06 \\

\textit{w/} Handcrafted RL Set  & 0.815 & 0.518 & 46.35 & 87.60 & 0.758 & \underline{68.61} \\

\textit{w/} APT & 0.859 & 0.523 & 50.73 & 87.88 & 0.795 & \textbf{71.26} \\
\midrule
\multicolumn{7}{l}{\emph{\textbf{SD3.5-Medium}}} \\
Base Model & 0.684 & 0.408 & 17.63 & 83.67 & 0.312 & 48.34 \\

\textit{w/} Handcrafted RL Set & 0.722 & 0.399 & 27.64 & 86.54 & 0.534 & \underline{55.94} \\

\textit{w/} APT & 0.763 & 0.404 & 29.50 & 86.50 & 0.569 & \textbf{57.92} \\
\midrule
\multicolumn{7}{l}{\emph{\textbf{FLUX.1 [Dev]}}} \\
Base Model & 0.660 & 0.388 & 22.23 & 82.54 & 0.260 & 47.11 \\

\textit{w/} Handcrafted RL Set & 0.721 & 0.435 & 29.47 & 85.53 & 0.310 & \underline{52.32} \\

\textit{w/} APT & 0.751 & 0.459 & 31.01 & 85.75 & 0.513 & \textbf{57.81} \\
\midrule
\multicolumn{7}{l}{\emph{\textbf{LongCat-Image}}} \\
Base Model & 0.727 & 0.500 & 28.60 & 85.71 & 0.426 & 55.92 \\

\textit{w/} Handcrafted RL Set & 0.780 & 0.503 & 31.14 & 87.80 & 0.730 & \underline{64.05} \\

\textit{w/} APT & 0.832 & 0.510 & 34.43 & 87.89 & 0.821 & \textbf{67.72} \\
\bottomrule
\end{tabular}%
}
\vspace{-3mm}
\end{table*}

\textbf{Base T2I Models.}
To demonstrate that APT is model-agnostic, we apply it to four diffusion-based T2I models: \textit{Lens}~\citep{t2i-lens}, \textit{SD3.5-Medium}~\citep{t2i-sd3.5-m}, \textit{FLUX.1 [dev]}~\citep{t2i-flux1.dev}, and \textit{LongCat-Image}~\citep{t2i-longcat-image}. We further evaluate whether APT generalizes across model architectures by applying it to the autoregressive image generator \textit{Janus-Pro}~\citep{t2i-janus-pro}. We use the same APT agentic workflow across all models, without introducing any model-specific prompt sets, reward designs, or hyperparameter searches.

\textbf{Benchmarks and Metrics.}
We evaluate on five widely used text-to-image benchmarks that jointly cover compositional generation, prompt following, and text rendering:
GenEval~\citep{bench-geneval}, GenEval2~\citep{bench-geneval2}, OneIG-EN~\citep{bench-oneig}, DPG~\citep{bench-dpg}, and CVTG-2K~\citep{bench-cvtg}. Details of each benchmark are provided in Section~\ref{sec:benchmark-details}.
Since the benchmark scores are reported on different scales, we additionally report a normalized average by rescaling GenEval, OneIG-EN, and CVTG-2K to the $[0,100]$ range and then averaging the scores across all five benchmarks.
Crucially, \textbf{none of these benchmarks is used to construct the training prompts}: a single APT run produces one post-trained model, which is then evaluated on all five benchmarks, directly testing the benchmark-agnostic nature of APT described in Section~\ref{sec:training-overall}.

\textbf{Implementation Details.}
We use DiffusionNFT~\citep{diffusionnft} as our default RL algorithm for post-training. All models are post-trained with LoRA (rank $64$, $\alpha=128$), using a learning rate of $3\mathrm{e}\text{-}4$.
During rollout, Lens, SD3.5-M, and LongCat-Image use $10$ sampling steps, while FLUX.1 [dev] uses $6$ steps. 
The training resolution is $768\times768$ for Lens and SD3.5-M, and $512\times512$ for FLUX.1 [dev] and LongCat-Image. For APT, each training round consists of $20$ optimization steps. At the beginning of each round, the Diagnosis Agent samples $120$ prompt-rubrics pairs from the evaluation set generated by the Evaluation Set Generation Agent and produces an evaluation report. This report is then fed into the RL Data Generation Agent, which automatically generates $480$ RL training prompts for the subsequent training round. For the handcrafted RL baseline, we use the same total number of training prompts as APT and follow the same rubric-generation and reward-evaluation procedures. Unless otherwise stated, FLUX.1 [dev] is post-trained for $200$ optimization steps, while all other models are post-trained for $400$ steps. At inference time, all models are evaluated at a resolution of $768\times768$ with the following settings: Lens ($20$ steps, CFG $=5.0$), SD3.5-M ($20$ steps, CFG $=4.5$), FLUX.1 [dev] ($16$ steps, CFG $=3.5$), and LongCat-Image ($20$ steps, CFG $=4.5$).

\subsection{Main Results}
\label{sec:exp-main}
In Table~\ref{tab:main_all}, we apply APT to four different base models and compare each APT-trained model against two counterparts: its original base model and a version post-trained on a recent \emph{handcrafted} RL training set proposed by Lens~\citep{t2i-lens} and manually curated by human experts. Importantly, all APT experiments use the same agentic workflow, without any model-specific prompt sets, reward designs, or hyperparameter tuning. We draw two main conclusions:

\begin{itemize}[leftmargin=*, labelsep=0.5em]
\item \textbf{APT Consistently Outperforms a Carefully Handcrafted RL Training Set Across All Four Base Models.}
The handcrafted RL set is carefully curated by human experts to cover a broad range of generation scenarios, making it a strong post-training baseline. Nevertheless, APT consistently achieves higher average performance across all four models. This suggests that manually designing RL training sets may not be necessary: human experts may struggle to anticipate all relevant failure modes or fully characterize the model-specific output distribution, whereas APT can automatically diagnose each model and adapt its training data accordingly.

\item \textbf{A Single APT Run Yields Consistent Improvements Across Multiple Benchmarks.}
Without using benchmark-specific training prompts or tuning, the same post-trained model improves over its base model across most evaluated benchmarks. This demonstrates strong benchmark-agnostic generalization, showing that APT does not rely on optimization toward any particular benchmark.
\end{itemize}

\begin{table*}[!t]
\vspace{-2mm}
\centering
\caption{\textbf{APT Captures Model-Specific Training Distributions.} Each row corresponds to a target model post-trained with RL data generated by APT from different source models. {\setlength{\fboxsep}{0pt}%
\smash{\colorbox{cyan!20}{\strut Diagonal}}} entries denote the standard APT setting, where the RL data is adaptively generated for the same target model, while off-diagonal entries reuse data generated for another model. Detailed results for each benchmark are shown in Tables~\ref{tab:transfer_geneval_oneig} in the Appendix~\ref{sec:benchmark-details}.}
\vspace{1mm}
\label{tab:transfer}
\small
\setlength{\tabcolsep}{5.0pt}
\renewcommand{\arraystretch}{1.06}

\begin{minipage}[t]{0.49\textwidth}
\centering
\textbf{(a) GenEval} \\
\vspace{1mm}
\resizebox{\linewidth}{!}{%
\begin{tabular}{lcccc}
\toprule
\textbf{Target/Source} & \textbf{Lens} & \textbf{SD3.5} & \textbf{FLUX} & \textbf{LongCat} \\
\midrule
Lens & \cellcolor{cyan!20}\textbf{0.851} & 0.833 & 0.846 & 0.798 \\
SD3.5 & 0.736 & \cellcolor{cyan!20}\textbf{0.743} & 0.742 & 0.722 \\
FLUX & 0.679 & 0.726 & \cellcolor{cyan!20}\textbf{0.751} & 0.729 \\
LongCat & 0.823 & 0.786 & 0.827 & \cellcolor{cyan!20}\textbf{0.828} \\
\bottomrule
\end{tabular}%
}
\end{minipage}
\hfill
\begin{minipage}[t]{0.49\textwidth}
\centering
\textbf{(b) OneIG-EN} \\
\vspace{1mm}
\resizebox{\linewidth}{!}{%
\begin{tabular}{lcccc}
\toprule
\textbf{Target/Source} & \textbf{Lens} & \textbf{SD3.5} & \textbf{FLUX} & \textbf{LongCat} \\
\midrule
Lens & \cellcolor{cyan!20}\textbf{0.523} & 0.522 & 0.517 & 0.520 \\
SD3.5 & 0.404 & \cellcolor{cyan!20}0.404 & \textbf{0.405} & 0.395 \\
FLUX & 0.436 & 0.443 & \cellcolor{cyan!20}\textbf{0.459} & 0.440 \\
LongCat & \textbf{0.509} & 0.496 & 0.508 & \cellcolor{cyan!20}0.506 \\
\bottomrule
\end{tabular}%
}
\end{minipage}

\vspace{3mm}

\begin{minipage}[t]{0.49\textwidth}
\centering
\textbf{(c) GenEval2} \\
\vspace{1mm}
\resizebox{\linewidth}{!}{%
\begin{tabular}{lcccc}
\toprule
\textbf{Target/Source} & \textbf{Lens} & \textbf{SD3.5} & \textbf{FLUX} & \textbf{LongCat} \\
\midrule
Lens & \cellcolor{cyan!20}\textbf{46.80} & 44.59 & 46.35 & 44.98 \\
SD3.5 & 23.61 & \cellcolor{cyan!20}\textbf{27.58} & 23.74 & 22.65 \\
FLUX & 30.06 & 28.69 & \cellcolor{cyan!20}\textbf{31.01} & 30.02 \\
LongCat & 34.91 & 37.37 & 34.90 & \cellcolor{cyan!20}\textbf{37.80} \\
\bottomrule
\end{tabular}%
}
\end{minipage}
\hfill
\begin{minipage}[t]{0.49\textwidth}
\centering
\textbf{(d) Average Over Three Benchmarks} \\
\vspace{1mm}
\resizebox{\linewidth}{!}{%
\begin{tabular}{lcccc}
\toprule
\textbf{Target/Source} & \textbf{Lens} & \textbf{SD3.5} & \textbf{FLUX} & \textbf{LongCat} \\
\midrule
Lens & \cellcolor{cyan!20}\textbf{61.40} & 60.03 & 60.88 & 58.93 \\
SD3.5 & 45.87 & \cellcolor{cyan!20}\textbf{47.43} & 46.15 & 44.78 \\
FLUX & 47.19 & 48.53 & \cellcolor{cyan!20}\textbf{50.67} & 48.97 \\
LongCat & 56.04 & 55.19 & 56.13 & \cellcolor{cyan!20}\textbf{57.07} \\
\bottomrule
\end{tabular}%
}
\end{minipage}
\vspace{-2mm}
\end{table*}

\subsection{Ablation Study}
\label{sec:exp-ablation}

\textbf{APT Captures Model-Specific Training Distributions.}
We investigate whether different T2I models exhibit distinct distributions, and whether APT can automatically adapt its RL data generation to each model. For each of the four models, we compare the standard APT setting against RL using the training data accumulated by APT when post-training another model. As shown in Table~\ref{tab:transfer}, each row corresponds to a target model being post-trained with RL data generated for different source models, while the diagonal entry represents the standard APT setting, where the RL data is generated adaptively for the same target model. 
The diagonal setting achieves the best average performance within every row across all four models. These results indicate that different T2I models exhibit distinct distributions, making RL data generated for one model suboptimal for another. APT addresses this issue by continuously diagnosing the current model and automatically generating RL data tailored to its evolving distribution, rather than relying on a fixed training set shared across models. Note that, for efficiency, we post-train each model for 200 steps in this study.

\begin{table*}[!t]
\vspace{-2mm}
\centering
\caption{\textbf{Ablation of the evaluation set generation agent.} We compare the APT-generated evaluation set with existing benchmark evaluation sets for producing diagnosis reports during post-training. Detailed results for each benchmark are shown in Tables~\ref{tab:eval_data_geneval_oneig}--\ref{tab:eval_data_dpg_cvtg} in the Appendix~\ref{sec:benchmark-details}.}
\vspace{1mm}
\label{tab:eval_data}
\small
\renewcommand{\arraystretch}{1.1}
\resizebox{\textwidth}{!}{%
\begin{tabular}{lcccccc}
\toprule
\multirow{2}{*}[-1pt]{\textbf{Model}} & \textbf{GenEval} & \textbf{OneIG-EN} & \textbf{GenEval2} & \textbf{DPG} & \textbf{CVTG-2K} & \multirow{2}{*}[-1pt]{\textbf{Average}} \\
\cmidrule(lr){2-6}
& Overall & Average & Soft-TIFA$_{\mathrm{GM}}$ & Overall & Word Accuracy & \\
\midrule

\multicolumn{7}{l}{\emph{\textbf{Lens}}} \\
Base Model & 0.702 & 0.521 & 29.06 & 85.93 & 0.480 & 57.06 \\

OneIG-EN Evaluation Set & 0.847 & 0.528 & 47.08 & 87.39 & 0.823 & \underline{70.85} \\

GenEval Evaluation Set & 0.887 & 0.505 & 41.43 & 84.23 & 0.674 & 66.45 \\

APT Evaluation Set & 0.859 & 0.523 & 50.73 & 87.88 & 0.795 & \textbf{71.26} \\

\midrule

\multicolumn{7}{l}{\emph{\textbf{SD3.5-Medium}}} \\
Base Model & 0.684 & 0.408 & 17.63 & 83.67 & 0.312 & 48.34 \\

OneIG-EN Evaluation Set & 0.800 & 0.431 & 26.38 & 86.69 & 0.497 & \underline{57.17} \\

GenEval Evaluation Set & 0.692 & 0.390 & 17.35 & 79.02 & 0.458 & 50.07 \\

APT Evaluation Set & 0.763 & 0.404 & 29.50 & 86.50 & 0.569 & \textbf{57.92} \\
\bottomrule
\end{tabular}%
}
\vspace{-5mm}
\end{table*}

\textbf{Ablation on the Evaluation Set Generation Agent.}
We study the role of the Evaluation Set Generation Agent by replacing the automatically generated APT evaluation set with existing benchmark evaluation sets when producing the diagnosis report. Specifically, we use the OneIG-EN evaluation set and the GenEval evaluation set as alternatives, while keeping the remaining APT pipeline and training configuration unchanged. As shown in Table~\ref{tab:eval_data}, using the APT-generated evaluation set consistently yields the best average performance on both Lens and SD3.5-Medium. In contrast, relying on a single existing benchmark tends to \textit{bias} the diagnosis toward the corresponding capability and leads to less balanced improvements across benchmarks. This result highlights the importance of automatically constructing a diverse and comprehensive evaluation set for reliable diagnosis and adaptive post-training.

\begin{table*}[!t]
\vspace{-2mm}
\centering
\caption{\textbf{Ablation on the Iterative Feedback Loop and the RL Data Generation Agent.} This study is conducted using the Lens base model. Detailed results for each benchmark are shown in Tables~\ref{tab:ablation_geneval_oneig}--\ref{tab:ablation_dpg_cvtg} in the Appendix~\ref{sec:benchmark-details}.}
\vspace{1mm}
\label{tab:ablation}
\small
\renewcommand{\arraystretch}{1.1}
\resizebox{\textwidth}{!}{%
\begin{tabular}{lcccccc}
\toprule
\multirow{2}{*}[-1pt]{\textbf{Setting}} & \textbf{GenEval} & \textbf{OneIG-EN} & \textbf{GenEval2} & \textbf{DPG} & \textbf{CVTG-2K} & \multirow{2}{*}[-1pt]{\textbf{Average}} \\
\cmidrule(lr){2-6}
& Overall & Average & Soft-TIFA$_{\mathrm{GM}}$ & Overall & Word Accuracy & \\
\midrule
APT & \textbf{0.859} & 0.523 & \textbf{50.73} & \textbf{87.88} & \textbf{0.795} & \textbf{71.26} \\
\midrule
\multicolumn{7}{l}{\emph{\textbf{Ablation on the Iterative Feedback Loop}}} \\
\textit{w/o} Iterative Feedback Loop & 0.844 & 0.523 & 48.27 & 87.61 & 0.791 & 70.34 \\
\midrule
\multicolumn{7}{l}{\emph{\textbf{Ablation on the RL Data Generation Agent}}} \\
Reduced Diagnostic Criteria & 0.800 & 0.520 & 49.58 & 87.53 & 0.763 & 69.08 \\
Hard-Only Sampling & 0.853 & \textbf{0.530} & 47.91 & 87.58 & 0.794 & 70.64 \\
\bottomrule
\end{tabular}%
}
\end{table*}

\textbf{Ablation on the Iterative Feedback Loop.}
We study whether repeatedly refreshing model feedback is necessary for effective post-training.
We compare APT with a variant that continues generating new training prompts but does not re-diagnose the updated model after each round, while keeping the total number of training prompts and RL optimization steps unchanged.
As shown in Table~\ref{tab:ablation}, removing the iterative feedback loop consistently degrades performance, demonstrating the benefit of continuously adapting the training distribution to the evolving model.

\textbf{Ablation on the RL Data Generation Agent.}
We further ablate the prompt-sampling strategy of the RL Data Generation Agent.
Removing part of the diagnostic criteria leads to a clear performance drop, while sampling only from the most difficult capabilities is also inferior to the full strategy.
These results show that effective RL data generation benefits from both comprehensive diagnostic feedback and a balanced distribution that prioritizes weak capabilities while retaining moderate and mastered ones.
The results are shown in Table~\ref{tab:ablation}, and the detailed instruction prompts for the two ablation settings are provided in Appendix~\ref{sec:prompt-agent-ablation-1} and Appendix~\ref{sec:prompt-agent-ablation-2}.

\textbf{Architecture Generalization to Autoregressive Generation.}
We further evaluate APT on an autoregressive image generation model, \textit{Janus-Pro}~\citep{t2i-janus-pro}, using GRPO~\citep{grpo} for RL post-training. As shown in Table~\ref{tab:ar_geneval_oneig_geneval2}, APT consistently outperforms both the original base model and the handcrafted RL post-training baseline across the evaluated benchmarks. These results demonstrate that APT is not tied to diffusion-based generators and can generalize effectively to autoregressive generation architectures, highlighting its architecture-agnostic nature.

\textbf{Robustness to Different LLM Backbones.}
We further examine the sensitivity of APT to the LLM backbone used by the RL Data Generation Agent. All tested backbones yield substantial gains in average performance on \textit{Lens}, suggesting that APT is robust across the evaluated LLM families and scales. Detailed results are provided in Appendix~\ref{sec:more_ablation}, Table~\ref{tab:ablation_llm}.

\begin{table*}[!t]
\vspace{-3mm}
\centering
\caption{\textbf{Architecture Generalization to Autoregressive Generation.} Comparison of APT with the original base model and a handcrafted RL post-training baseline on the AR model \textbf{Janus-Pro}.}
\vspace{1mm}
\label{tab:ar_geneval_oneig_geneval2}
\scriptsize
\setlength{\tabcolsep}{2.8pt}
\renewcommand{\arraystretch}{1.1}
\resizebox{\textwidth}{!}{%
\begin{tabular}{lccccccc|cccccc|c}
\toprule
\multirow{2}{*}[-1pt]{\textbf{Model}} & \multicolumn{7}{c|}{\textbf{GenEval}} & \multicolumn{6}{c|}{\textbf{OneIG-EN}} & \multicolumn{1}{c}{\textbf{GenEval2}} \\
\cmidrule(lr){2-8} \cmidrule(lr){9-14} \cmidrule(lr){15-15}
& Single & Two & Count & Color & Position & Attribute & \textbf{Overall} & Alignment & Text & Reasoning & Style & Diversity & \textbf{Average} & \textbf{Soft-TIFA}$\bm{_{\mathrm{GM}}}$ \\
\midrule
\multicolumn{7}{l}{\emph{\textbf{Janus-Pro}}} \\
Base Model 
& 0.994 & 0.869 & 0.538 & 0.891 & 0.758 & 0.635 & 0.781 & 0.799 & 0.068 & 0.165 & 0.293 & 0.243 & 0.314 & 15.25 \\
\textit{w/} Handcrafted RL Set 
& 0.984 & 0.891 & 0.609 & 0.902 & 0.790 & 0.660 & \underline{0.806} & 0.804 & 0.074 & 0.175 & 0.297 & 0.229 & \underline{0.316} & \underline{20.42} \\
\textit{w/} APT
& 0.991 & 0.909 & 0.606 & 0.910 & 0.808 & 0.668 & \textbf{0.815} & 0.809 & 0.074 & 0.175 & 0.297 & 0.233 & \textbf{0.318} & \textbf{21.61} \\
\bottomrule
\end{tabular}%
}
\vspace{-4mm}
\end{table*}


%% file: sections/5-conclusion.tex
\section{Conclusion}
\label{sec:conclusion}
We presented APT, a fully automated framework for adaptive post-training of text-to-image models. By coordinating evaluation set generation, model diagnosis, and RL data generation, APT adapts the training distribution to each model's evolving capabilities. Experiments across multiple models and benchmarks show higher average performance than post-training on handcrafted RL data, while transfer studies and ablations support the value of model-specific training data and iterative feedback. These findings highlight feedback-driven task adaptation as a practical direction for automated post-training and a step toward recursive self-improvement in text-to-image generation.

%% file: sections/X-appendix.tex
\newpage
\appendix

\section{Related Work}
\label{sec:related_work}

\textbf{Post-Training for Text-to-Image Generation.} 
Post-training has become a standard paradigm for adapting pre-trained T2I models to downstream objectives. Supervised fine-tuning (SFT) adapts a model to curated image--text pairs, with performance gains largely determined by data filtering, recaptioning, and aesthetic curation, but it remains bounded by the collected corpus and costly human curation. A complementary line directly optimizes downstream rewards following the policy-gradient view of reinforcement learning~\citep{williams1992simple, schulman2017proximal}: DDPO~\citep{black2024training} formulates denoising as a multi-step decision-making problem and optimizes diffusion models with policy gradients; Diffusion-DPO~\citep{wallace2024diffusion} adapts direct preference optimization~\citep{rafailov2023direct} to diffusion models, and SPO~\citep{liang2024step} refines it with step-aware preference signals; more recently, Flow-GRPO~\citep{flow-grpo} and follow-ups~\citep{li2025mixgrpo, liu2026policy} extend online policy-gradient training to flow matching via stochastic sampling, and DiffusionNFT~\citep{diffusionnft} performs negative-aware fine-tuning on the forward process without likelihood estimation.

\textbf{Agentic Image Generation.}
Recent work augments image generation and editing with agentic capabilities such as planning~\citep{agent-plan}, reasoning over implicit user intent~\citep{agent-reason}, web and image search for open-world grounding~\citep{agent-search}, and self-feedback for test-time scaling~\citep{agent-feedback-1,agent-feedback-2}, with emerging systems unifying these abilities in a single framework~\citep{agent-qwen}. 

These methods share a common design: \emph{the agent operates at inference time around a frozen generator}, improving how a fixed model is used rather than changing the model itself. As a result, intrinsic failure modes persist once the agentic scaffolding is removed, and every deployment pays the extra inference cost. 
APT instead moves the agent from the \textit{inference} loop into the \emph{post-training} loop: discover model weaknesses through evaluation and address them through training. The gains are absorbed into the model parameters and retained under standard single-pass inference, and the training distribution is re-derived from the current model at every round, making APT \emph{model-adaptive} rather than applying one fixed agentic policy to any generator. The two directions are thus complementary: prior work makes better use of a given model, while APT makes the model itself better.

\section{More Ablation Studies}
\label{sec:more_ablation}

\textbf{Robustness to Different LLM Backbones in the RL Data Generation Agent.}
We investigate how sensitive APT is to the choice of LLM backbone used by the RL Data Generation Agent. Using \textit{Lens} as the base image generation model, we replace the default GPT-5.4 backbone with LLMs from different families and model scales, while keeping all other components of the post-training pipeline fixed. As shown in Table~\ref{tab:ablation_llm}, all tested LLM backbones lead to effective post-training and achieve comparable improvements across the five benchmarks, despite noticeable differences in their absolute performance. This suggests that the effectiveness of APT does not depend on a specific LLM family or scale, and that the RL Data Generation Agent remains robust across diverse proprietary and open-source backbones.

\begin{table*}[h]
\centering
\caption{Ablation on the LLM backbone used in the RL Data Generation Agent. We replace the default GPT-5.4 with LLMs from different families and scales while keeping the remaining APT pipeline unchanged. Results are reported on five benchmarks using \textit{Lens} as the base model.}
\vspace{1mm}
\label{tab:ablation_llm}
\small
\setlength{\tabcolsep}{7.0pt}
\renewcommand{\arraystretch}{1.12}
\resizebox{\textwidth}{!}{%
\begin{tabular}{lcccccc}
\toprule
\multirow{2}{*}[-1pt]{\textbf{Model}} & \textbf{GenEval} & \textbf{OneIG-EN} & \textbf{GenEval2} & \textbf{DPG} & \textbf{CVTG-2K} & \multirow{2}{*}[-1pt]{\textbf{Average}} \\
\cmidrule(lr){2-6}
& Overall & Average & Soft-TIFA$_{\mathrm{GM}}$ & Overall & Word Accuracy & \\
\midrule
Base Model & 0.702 & 0.521 & 29.06 & 85.93 & 0.480 & 57.06 \\
\midrule
\multicolumn{7}{l}{\emph{\textbf{GPT Series}}} \\
GPT-5.4
& 0.859 & 0.523 & 50.73 & 87.88 & 0.795 & 71.26 \\
GPT-5.6-Sol
& 0.832 & 0.519 & 51.43 & 88.12 & 0.791 & 70.75 \\
GPT-4.1
& 0.851 & 0.520 & 48.94 & 87.93 & 0.796 & 70.71 \\
GPT-5.4-mini
& 0.825 & 0.518 & 50.79 & 87.69 & 0.810 & 70.76 \\
\midrule
\multicolumn{7}{l}{\emph{\textbf{Qwen Series}}} \\
Qwen3.5-9B
& 0.861 & 0.520 & 47.03 & 88.36 & 0.793 & 70.56 \\
Qwen3.5-122B-A10B
& 0.893 & 0.525 & 51.42 & 87.65 & 0.805 & 72.27 \\
\midrule
\multicolumn{7}{l}{\emph{\textbf{Gemini Series}}} \\
Gemini-3.8-Flash
& 0.829 & 0.525 & 48.67 & 87.70 & 0.781 & 69.97 \\
\midrule
\multicolumn{7}{l}{\emph{\textbf{Grok Series}}} \\
Grok-4.6
& 0.819 & 0.501 & 51.44 & 87.50 & 0.732 & 68.83 \\
\bottomrule
\end{tabular}%
}
\end{table*}

\section{Detailed Benchmark Results}
\label{sec:benchmark-details}

We evaluate APT on five benchmarks, with the details of each benchmark described as follows:
\begin{itemize}
    \item \textbf{GenEval}~\citep{bench-geneval} focuses on object-centric compositional alignment, with 553 prompts spanning six templated tasks: single-object generation, two-object co-occurrence, counting, color, spatial relation, and attribute binding. It uses detector- and classifier-based verification to assess whether generated images satisfy these structured constraints.
    \item \textbf{OneIG-EN}~\citep{bench-oneig} is the English split of OneIG-Bench, comprising 1,120 prompts across general objects, portraits, anime/stylization, text rendering, and knowledge/reasoning. It evaluates generated images with dimension-specific scores for subject alignment, text accuracy, reasoning, style, diversity, and overall performance.
    \item \textbf{GenEval2}~\citep{bench-geneval2} extends GenEval with 800 prompts of increasing compositionality, covering 40 objects, 18 attributes, 9 relations, and counts from 2 to 7. Each prompt contains 3 to 10 visual atoms, including objects, attributes, counts, and relations, and is evaluated using Soft-TIFA, which decomposes prompts into atom-level visual questions and aggregates soft correctness scores to measure fine-grained compositional alignment. Soft-TIFA$_{\mathrm{AM}}$ averages these scores arithmetically within each image to estimate atom-level correctness, whereas Soft-TIFA$_{\mathrm{GM}}$ uses their geometric mean to capture prompt-level correctness by more strongly penalizing poorly satisfied atoms. Both variants then average the image-level scores arithmetically across the benchmark. We report Soft-TIFA$_{\mathrm{GM}}$, as the GenEval2~\citep{bench-geneval2} finds that it achieves higher AUROC against prompt-level human judgments than Soft-TIFA$_{\mathrm{AM}}$ across the original, rewritten, and combined prompt sets.
    \item \textbf{DPG}~\citep{bench-dpg} evaluates dense prompt following with 1,065 lengthy prompts describing multiple objects, fine-grained attributes, and complex relationships. It decomposes each prompt into a dependency-aware question graph and employs an MLLM-based VQA evaluator to measure alignment across global, entity, attribute, relation, and other semantic dimensions.
    \item \textbf{CVTG-2K}~\citep{bench-cvtg} evaluates complex visual text generation with multiple text regions, where prompts vary the number of regions from 2 to 5 and specify text content, position, length, and style attributes such as color, font, and size.
\end{itemize}

Table~\ref{tab:main_geneval} provides the detailed \textbf{GenEval} results corresponding to Table~\ref{tab:main_all}.

Table~\ref{tab:main_oneig_en} provides the detailed \textbf{OneIG-EN} results corresponding to Table~\ref{tab:main_all}. We note that RL post-training can substantially reduce generation diversity, a phenomenon observed for both the Lens~\citep{t2i-lens} and Mage-Flow~\citep{t2i-mage} models. Consequently, the \textbf{Diversity} score on OneIG-EN decreases markedly after RL post-training and may offset improvements in other dimensions, occasionally resulting in an average score lower than that of the base model. Nevertheless, the post-trained models consistently exhibit stronger performance in capability-oriented dimensions such as subject alignment, text rendering, and reasoning. More importantly, despite the diversity degradation shared by RL-based approaches, our method \textit{w/} APT generally achieves higher alignment, text, reasoning, and overall scores than \textit{w/} Handcrafted RL Set. These results suggest that APT produces more effective post-training data and improves the target capabilities more effectively than handcrafted RL post-training.

Table~\ref{tab:main_dpg} provides the detailed \textbf{DPG} results corresponding to Table~\ref{tab:main_all}.

Table~\ref{tab:main_cvtg2k} provides the detailed \textbf{CVTG-2K} results corresponding to Table~\ref{tab:main_all}.

Table~\ref{tab:transfer_geneval_oneig} provides the detailed \textbf{GenEval} and \textbf{OneIG-EN} results corresponding to Table~\ref{tab:transfer}.

Table~\ref{tab:eval_data_geneval_oneig} provides the detailed \textbf{GenEval} and \textbf{OneIG-EN} results corresponding to Table~\ref{tab:eval_data}.

Table~\ref{tab:eval_data_dpg_cvtg} provides the detailed \textbf{DPG} and \textbf{CVTG-2K} results corresponding to Table~\ref{tab:eval_data}.

Table~\ref{tab:ablation_geneval_oneig} provides the detailed \textbf{GenEval} and \textbf{OneIG-EN} results corresponding to Table~\ref{tab:ablation}.

Table~\ref{tab:ablation_dpg_cvtg} provides the detailed \textbf{DPG} and \textbf{CVTG-2K} results corresponding to Table~\ref{tab:ablation}.


\begin{table*}[h]
\centering
\caption{Detailed comparison of APT with the original base models and a handcrafted RL post-training
baseline across four T2I models on \textbf{GenEval}.}
\vspace{1mm}
\label{tab:main_geneval}
\scriptsize
\setlength{\tabcolsep}{4.5pt}
\renewcommand{\arraystretch}{1.12}
\resizebox{\textwidth}{!}{%
\begin{tabular}{lccccccc}
\toprule
\textbf{Model}
& Single & Two & Count & Color & Position & Attribute & \textbf{Overall} \\
\midrule
\multicolumn{8}{l}{\emph{\textbf{Lens}}} \\
Base Model
& 0.991 & 0.788 & 0.547 & 0.848 & 0.548 & 0.488 & 0.702 \\
\textit{w/} Handcrafted RL Set
& 0.994 & 0.929 & 0.763 & 0.848 & 0.735 & 0.620 & \underline{0.815} \\
\textit{w/} APT
& 1.000 & 0.947 & 0.822 & 0.872 & 0.835 & 0.680 & \textbf{0.859} \\
\midrule
\multicolumn{8}{l}{\emph{\textbf{SD3.5-Medium}}} \\
Base Model
& 0.994 & 0.854 & 0.593 & 0.844 & 0.260 & 0.560 & 0.684 \\
\textit{w/} Handcrafted RL Set
& 0.997 & 0.955 & 0.622 & 0.840 & 0.370 & 0.550 & \underline{0.722} \\
\textit{w/} APT
& 0.997 & 0.932 & 0.794 & 0.862 & 0.335 & 0.660 & \textbf{0.763} \\
\midrule
\multicolumn{8}{l}{\emph{\textbf{FLUX.1 [Dev]}}} \\
Base Model
& 0.981 & 0.813 & 0.709 & 0.801 & 0.228 & 0.428 & 0.660 \\
\textit{w/} Handcrafted RL Set
& 0.972 & 0.869 & 0.813 & 0.798 & 0.295 & 0.580 & \underline{0.721} \\
\textit{w/} APT
& 0.988 & 0.924 & 0.756 & 0.846 & 0.353 & 0.638 & \textbf{0.751} \\
\midrule
\multicolumn{8}{l}{\emph{\textbf{LongCat-Image}}} \\
Base Model
& 0.975 & 0.854 & 0.575 & 0.846 & 0.528 & 0.583 & 0.727 \\
\textit{w/} Handcrafted RL Set
& 0.984 & 0.939 & 0.569 & 0.854 & 0.730 & 0.605 & \underline{0.780} \\
\textit{w/} APT
& 0.984 & 0.960 & 0.656 & 0.899 & 0.780 & 0.710 & \textbf{0.832} \\
\bottomrule
\end{tabular}%
}
\end{table*}

\begin{table*}[h]
\centering
\caption{Detailed comparison of APT with the original base models and a handcrafted RL post-training
baseline across four T2I models on \textbf{OneIG-EN}.}
\vspace{1mm}
\label{tab:main_oneig_en}
\scriptsize
\setlength{\tabcolsep}{5.0pt}
\renewcommand{\arraystretch}{1.12}
\resizebox{\textwidth}{!}{%
\begin{tabular}{lcccccc}
\toprule
\textbf{Model}
& Alignment & Text & Reasoning & Style & Diversity & \textbf{Average} \\
\midrule
\multicolumn{7}{l}{\emph{\textbf{Lens}}} \\
Base Model
& 0.840 & 0.872 & 0.247 & 0.387 & 0.257 & \underline{0.521} \\
\textit{w/} Handcrafted RL Set
& 0.868 & 0.949 & 0.274 & 0.314 & 0.184 & 0.518 \\
\textit{w/} APT
& 0.874 & 0.947 & 0.284 & 0.335 & 0.174 & \textbf{0.523} \\
\midrule
\multicolumn{7}{l}{\emph{\textbf{SD3.5-Medium}}} \\
Base Model
& 0.793 & 0.341 & 0.288 & 0.363 & 0.254 & \textbf{0.408} \\
\textit{w/} Handcrafted RL Set
& 0.823 & 0.379 & 0.302 & 0.316 & 0.173 & 0.399 \\
\textit{w/} APT
& 0.825 & 0.410 & 0.305 & 0.314 & 0.164 & \underline{0.404} \\
\midrule
\multicolumn{7}{l}{\emph{\textbf{FLUX.1 [Dev]}}} \\
Base Model
& 0.780 & 0.332 & 0.234 & 0.352 & 0.244 & 0.388 \\
\textit{w/} Handcrafted RL Set
& 0.824 & 0.548 & 0.264 & 0.330 & 0.208 & \underline{0.435} \\
\textit{w/} APT
& 0.824 & 0.655 & 0.272 & 0.340 & 0.203 & \textbf{0.459} \\
\midrule
\multicolumn{7}{l}{\emph{\textbf{LongCat-Image}}} \\
Base Model
& 0.816 & 0.750 & 0.228 & 0.396 & 0.309 & 0.500 \\
\textit{w/} Handcrafted RL Set
& 0.863 & 0.877 & 0.235 & 0.322 & 0.220 & \underline{0.503} \\
\textit{w/} APT
& 0.872 & 0.907 & 0.236 & 0.322 & 0.215 & \textbf{0.510} \\
\bottomrule
\end{tabular}%
}
\end{table*}

\begin{table*}[h]
\centering
\caption{Detailed comparison of APT with the original base models and a handcrafted RL post-training
baseline across four T2I models on \textbf{DPG}.}
\vspace{1mm}
\label{tab:main_dpg}
\scriptsize
\setlength{\tabcolsep}{5.0pt}
\renewcommand{\arraystretch}{1.12}
\resizebox{\textwidth}{!}{%
\begin{tabular}{lcccccc}
\toprule
\textbf{Model}
& Global & Entity & Attribute & Relation & Other & \textbf{Overall} \\
\midrule
\multicolumn{7}{l}{\emph{\textbf{Lens}}} \\
Base Model
& 93.47 & 90.44 & 91.05 & 92.16 & 90.49 & 85.93 \\
\textit{w/} Handcrafted RL Set
& 89.88 & 92.69 & 91.77 & 93.98 & 91.90 & \underline{87.60} \\
\textit{w/} APT
& 93.46 & 92.94 & 92.73 & 92.46 & 91.73 & \textbf{87.88} \\
\midrule
\multicolumn{7}{l}{\emph{\textbf{SD3.5-Medium}}} \\
Base Model
& 92.15 & 89.67 & 88.60 & 91.89 & 85.19 & 83.67 \\
\textit{w/} Handcrafted RL Set
& 90.34 & 91.98 & 91.71 & 92.27 & 89.96 & \textbf{86.54} \\
\textit{w/} APT
& 91.53 & 93.05 & 90.82 & 91.50 & 90.91 & \underline{86.50} \\
\midrule
\multicolumn{7}{l}{\emph{\textbf{FLUX.1 [Dev]}}} \\
Base Model
& 87.53 & 88.41 & 88.21 & 89.55 & 88.75 & 82.54 \\
\textit{w/} Handcrafted RL Set
& 91.37 & 91.12 & 89.83 & 91.30 & 87.78 & \underline{85.53} \\
\textit{w/} APT
& 92.25 & 91.49 & 90.61 & 88.77 & 90.51 & \textbf{85.75} \\
\midrule
\multicolumn{7}{l}{\emph{\textbf{LongCat-Image}}} \\
Base Model
& 85.68 & 90.39 & 90.21 & 91.78 & 88.68 & 85.71 \\
\textit{w/} Handcrafted RL Set
& 91.76 & 93.11 & 92.24 & 92.41 & 93.10 & \underline{87.80} \\
\textit{w/} APT
& 93.11 & 93.10 & 92.17 & 91.59 & 92.84 & \textbf{87.89} \\
\bottomrule
\end{tabular}%
}
\end{table*}

\begin{table*}[h]
\centering
\caption{Detailed comparison of APT with the original base models and a handcrafted RL post-training
baseline across four T2I models on \textbf{CVTG-2K}.}
\vspace{1mm}
\label{tab:main_cvtg2k}
\scriptsize
\setlength{\tabcolsep}{5.0pt}
\renewcommand{\arraystretch}{1.12}
\resizebox{\textwidth}{!}{%
\begin{tabular}{lccccc}
\toprule
\textbf{Model}
& \textbf{Word Accuracy} & NED & CLIPScore & VQAScore & Aesthetic \\
\midrule
\multicolumn{6}{l}{\emph{\textbf{Lens}}} \\
Base Model
& 0.480 & 0.711 & 0.785 & 0.932 & 4.986 \\
\textit{w/} Handcrafted RL Set
& \underline{0.758} & 0.878 & 0.786 & 0.959 & 5.926 \\
\textit{w/} APT
& \textbf{0.795} & 0.905 & 0.784 & 0.963 & 5.999 \\
\midrule
\multicolumn{6}{l}{\emph{\textbf{SD3.5-Medium}}} \\
Base Model
& 0.312 & 0.626 & 0.754 & 0.891 & 5.541 \\
\textit{w/} Handcrafted RL Set
& \underline{0.534} & 0.798 & 0.783 & 0.943 & 5.825 \\
\textit{w/} APT
& \textbf{0.569} & 0.837 & 0.819 & 0.938 & 5.604 \\
\midrule
\multicolumn{6}{l}{\emph{\textbf{FLUX.1 [Dev]}}} \\
Base Model
& 0.260 & 0.499 & 0.710 & 0.858 & 6.050 \\
\textit{w/} Handcrafted RL Set
& \underline{0.310} & 0.557 & 0.742 & 0.915 & 6.053 \\
\textit{w/} APT
& \textbf{0.513} & 0.725 & 0.777 & 0.931 & 6.404 \\
\midrule
\multicolumn{6}{l}{\emph{\textbf{LongCat-Image}}} \\
Base Model
& 0.426 & 0.682 & 0.756 & 0.935 & 5.501 \\
\textit{w/} Handcrafted RL Set
& \underline{0.730} & 0.866 & 0.781 & 0.960 & 6.032 \\
\textit{w/} APT
& \textbf{0.821} & 0.927 & 0.775 & 0.961 & 6.054 \\
\bottomrule
\end{tabular}%
}
\end{table*}



\begin{table*}[h]
\centering
\caption{Detailed results of \textbf{APT Captures Model-Specific Training Distributions} on \textbf{GenEval} and \textbf{OneIG-EN}. Each block corresponds to a target model post-trained with RL data generated by APT from different source models, while rows denote the source model used for data generation. {\setlength{\fboxsep}{0pt}%
\smash{\colorbox{cyan!20}{\strut Shaded rows}}} denote the standard APT setting.}
\vspace{1mm}
\label{tab:transfer_geneval_oneig}
\scriptsize
\setlength{\tabcolsep}{2.6pt}
\renewcommand{\arraystretch}{1.10}
\resizebox{\textwidth}{!}{%
\begin{tabular}{lccccccc|cccccc}
\toprule
\multirow{2}{*}[-1pt]{\textbf{Data Source}}
& \multicolumn{7}{c|}{\textbf{GenEval}}
& \multicolumn{6}{c}{\textbf{OneIG-EN}} \\
\cmidrule(lr){2-8} \cmidrule(lr){9-14}
& Single & Two & Count & Color & Position & Attribute & \textbf{Overall}
& Alignment & Text & Reasoning & Style & Diversity & \textbf{Average} \\
\midrule
\multicolumn{14}{l}{\emph{\textbf{Target Model: Lens}}} \\
\rowcolor{cyan!20}
Lens & 1.000 & 0.960 & 0.831 & 0.872 & 0.768 & 0.675 & \textbf{0.851}
& 0.871 & 0.952 & 0.278 & 0.338 & 0.176 & \textbf{0.523} \\
SD3.5-Medium & 1.000 & 0.934 & 0.813 & 0.862 & 0.748 & 0.640 & 0.833
& 0.869 & 0.941 & 0.277 & 0.343 & 0.181 & 0.522 \\
FLUX.1 [Dev] & 0.997 & 0.939 & 0.797 & 0.875 & 0.813 & 0.658 & 0.846
& 0.871 & 0.950 & 0.276 & 0.317 & 0.173 & 0.517 \\
LongCat-Image & 0.997 & 0.917 & 0.728 & 0.819 & 0.733 & 0.595 & 0.798
& 0.865 & 0.946 & 0.270 & 0.321 & 0.199 & 0.520 \\
\midrule
\multicolumn{14}{l}{\emph{\textbf{Target Model: SD3.5-Medium}}} \\
Lens & 0.994 & 0.904 & 0.794 & 0.811 & 0.318 & 0.595 & 0.736
& 0.821 & 0.398 & 0.309 & 0.316 & 0.174 & 0.404 \\
\rowcolor{cyan!20}
SD3.5-Medium & 0.988 & 0.907 & 0.781 & 0.835 & 0.343 & 0.605 & \textbf{0.743}
& 0.824 & 0.402 & 0.307 & 0.322 & 0.166 & 0.404 \\
FLUX.1 [Dev] & 0.994 & 0.904 & 0.797 & 0.840 & 0.335 & 0.580 & 0.742
& 0.820 & 0.379 & 0.312 & 0.325 & 0.189 & \textbf{0.405} \\
LongCat-Image & 1.000 & 0.957 & 0.550 & 0.848 & 0.313 & 0.665 & 0.722
& 0.828 & 0.383 & 0.292 & 0.303 & 0.171 & 0.395 \\
\midrule
\multicolumn{14}{l}{\emph{\textbf{Target Model: FLUX.1 [Dev]}}} \\
Lens & 0.997 & 0.833 & 0.709 & 0.785 & 0.260 & 0.488 & 0.679
& 0.810 & 0.558 & 0.264 & 0.331 & 0.215 & 0.436 \\
SD3.5-Medium & 0.988 & 0.884 & 0.763 & 0.827 & 0.298 & 0.598 & 0.726
& 0.812 & 0.600 & 0.262 & 0.338 & 0.201 & 0.443 \\
\rowcolor{cyan!20}
FLUX.1 [Dev] & 0.988 & 0.924 & 0.756 & 0.846 & 0.353 & 0.638 & \textbf{0.751}
& 0.824 & 0.655 & 0.272 & 0.340 & 0.203 & \textbf{0.459} \\
LongCat-Image & 1.000 & 0.922 & 0.738 & 0.817 & 0.273 & 0.625 & 0.729
& 0.811 & 0.594 & 0.269 & 0.322 & 0.204 & 0.440 \\
\midrule
\multicolumn{14}{l}{\emph{\textbf{Target Model: LongCat-Image}}} \\
Lens & 0.994 & 0.970 & 0.644 & 0.880 & 0.738 & 0.710 & 0.823
& 0.862 & 0.907 & 0.235 & 0.316 & 0.224 & \textbf{0.509} \\
SD3.5-Medium & 0.997 & 0.955 & 0.653 & 0.891 & 0.603 & 0.620 & 0.786
& 0.860 & 0.865 & 0.234 & 0.309 & 0.211 & 0.496 \\
FLUX.1 [Dev] & 0.991 & 0.942 & 0.763 & 0.864 & 0.740 & 0.660 & 0.827
& 0.860 & 0.882 & 0.235 & 0.338 & 0.224 & 0.508 \\
\rowcolor{cyan!20}
LongCat-Image & 0.991 & 0.972 & 0.688 & 0.910 & 0.728 & 0.683 & \textbf{0.828}
& 0.863 & 0.880 & 0.237 & 0.324 & 0.226 & 0.506 \\
\bottomrule
\end{tabular}%
}
\end{table*}


\begin{table*}[h]
\centering
\caption{Detailed ablation of the evaluation set generation agent on \textbf{GenEval} and \textbf{OneIG-EN}.}
\vspace{1mm}
\label{tab:eval_data_geneval_oneig}
\scriptsize
\setlength{\tabcolsep}{3.2pt}
\renewcommand{\arraystretch}{1.12}
\resizebox{\textwidth}{!}{%
\begin{tabular}{lccccccc|cccccc}
\toprule
\multirow{2}{*}[-1pt]{\textbf{Model}} & \multicolumn{7}{c|}{\textbf{GenEval}} & \multicolumn{6}{c}{\textbf{OneIG-EN}} \\
\cmidrule(lr){2-8} \cmidrule(lr){9-14}
& Single & Two & Count & Color & Position & Attribute & \textbf{Overall} & Alignment & Text & Reasoning & Style & Diversity & \textbf{Average} \\
\midrule
\multicolumn{14}{l}{\emph{\textbf{Lens}}} \\
Base Model 
& 0.991 & 0.788 & 0.547 & 0.848 & 0.548 & 0.488 & 0.702 & 0.840 & 0.872 & 0.247 & 0.387 & 0.257 & 0.521 \\
OneIG-EN Evaluation Set 
& 0.997 & 0.929 & 0.788 & 0.912 & 0.770 & 0.688 & 0.847 & 0.869 & 0.969 & 0.292 & 0.344 & 0.166 & \textbf{0.528} \\
GenEval Evaluation Set 
& 0.984 & 0.957 & 0.900 & 0.864 & 0.930 & 0.685 & \textbf{0.887} & 0.823 & 0.908 & 0.273 & 0.330 & 0.193 & 0.505 \\
APT Evaluation Set 
& 1.000 & 0.947 & 0.822 & 0.872 & 0.835 & 0.680 & \underline{0.859} & 0.874 & 0.947 & 0.284 & 0.335 & 0.174 & \underline{0.523} \\
\midrule
\multicolumn{14}{l}{\emph{\textbf{SD3.5-Medium}}} \\
Base Model 
& 0.994 & 0.854 & 0.593 & 0.844 & 0.260 & 0.560 & 0.684 & 0.793 & 0.341 & 0.288 & 0.363 & 0.254 & \underline{0.408} \\
OneIG-EN Evaluation Set 
& 1.000 & 0.957 & 0.859 & 0.880 & 0.430 & 0.675 & \textbf{0.800} & 0.832 & 0.463 & 0.302 & 0.378 & 0.180 & \textbf{0.431} \\
GenEval Evaluation Set
& 0.959 & 0.745 & 0.663 & 0.745 & 0.680 & 0.363 & 0.692 & 0.693 & 0.413 & 0.287 & 0.330 & 0.227 & 0.390 \\
APT Evaluation Set 
& 0.997 & 0.932 & 0.794 & 0.862 & 0.335 & 0.660 & \underline{0.763} & 0.825 & 0.410 & 0.305 & 0.314 & 0.164 & 0.404 \\
\bottomrule
\end{tabular}%
}
\end{table*}

\begin{table*}[h]
\centering
\caption{Detailed ablation of the evaluation set generation agent on \textbf{DPG} and \textbf{CVTG-2K}.}
\vspace{1mm}
\label{tab:eval_data_dpg_cvtg}
\scriptsize
\setlength{\tabcolsep}{3.0pt}
\renewcommand{\arraystretch}{1.12}
\resizebox{\textwidth}{!}{%
\begin{tabular}{l|cccccc|ccccc}
\toprule
\multirow{2}{*}[-1pt]{\textbf{Model}}
& \multicolumn{6}{c|}{\textbf{DPG}}
& \multicolumn{5}{c}{\textbf{CVTG-2K}} \\
\cmidrule(lr){2-7} \cmidrule(lr){8-12}
& Global & Entity & Attribute & Relation & Other & \textbf{Overall}
& \textbf{Word Accuracy} & NED & CLIPScore & VQAScore & Aesthetic \\
\midrule
\multicolumn{12}{l}{\emph{\textbf{Lens}}} \\
Base Model
& 93.47 & 90.44 & 91.05 & 92.16 & 90.49 & 85.93
& 0.480 & 0.711 & 0.785 & 0.932 & 4.986 \\
OneIG-EN Evaluation Set
& 88.74 & 93.00 & 91.94 & 90.85 & 93.26 & \underline{87.39}
& \textbf{0.823} & 0.938 & 0.814 & 0.961 & 5.718 \\
GenEval Evaluation Set
& 88.46 & 91.78 & 89.81 & 87.63 & 88.77 & 84.23
& 0.674 & 0.852 & 0.797 & 0.941 & 5.182 \\
APT Evaluation Set
& 93.46 & 92.94 & 92.73 & 92.46 & 91.73 & \textbf{87.88}
& \underline{0.795} & 0.905 & 0.784 & 0.963 & 5.999 \\
\midrule
\multicolumn{12}{l}{\emph{\textbf{SD3.5-Medium}}} \\
Base Model
& 92.15 & 89.67 & 88.60 & 91.88 & 85.19 & 83.67
& 0.312 & 0.626 & 0.754 & 0.891 & 5.541 \\
OneIG-EN Evaluation Set
& 91.38 & 91.26 & 91.29 & 93.36 & 90.36 & \textbf{86.69}
& \underline{0.497} & 0.771 & 0.802 & 0.939 & 5.722 \\
GenEval Evaluation Set
& 86.54 & 85.01 & 86.68 & 84.52 & 86.99 & 79.02
& 0.458 & 0.770 & 0.780 & 0.882 & 4.797 \\
APT Evaluation Set
& 91.53 & 93.05 & 90.82 & 91.50 & 90.91 & \underline{86.50}
& \textbf{0.569} & 0.837 & 0.819 & 0.938 & 5.604 \\
\bottomrule
\end{tabular}%
}
\end{table*}



\begin{table*}[h]
\centering
\caption{Detailed ablation of the Iterative Feedback Loop and the RL Data Generation Agent on \textbf{GenEval} and \textbf{OneIG-EN}.}
\vspace{1mm}
\label{tab:ablation_geneval_oneig}
\scriptsize
\setlength{\tabcolsep}{3.2pt}
\renewcommand{\arraystretch}{1.12}
\resizebox{\textwidth}{!}{%
\begin{tabular}{lccccccc|cccccc}
\toprule
\multirow{2}{*}[-1pt]{\textbf{Setting}} & \multicolumn{7}{c|}{\textbf{GenEval}} & \multicolumn{6}{c}{\textbf{OneIG-EN}} \\
\cmidrule(lr){2-8} \cmidrule(lr){9-14}
& Single & Two & Count & Color & Position & Attribute & \textbf{Overall} & Alignment & Text & Reasoning & Style & Diversity & \textbf{Average} \\
\midrule
APT 
& 1.000 & 0.947 & 0.822 & 0.872 & 0.835 & 0.680 & \textbf{0.859} & 0.874 & 0.947 & 0.284 & 0.335 & 0.174 & 0.523 \\
\midrule
\multicolumn{14}{l}{\emph{\textbf{Ablation on the Iterative Feedback Loop}}} \\
\textit{w/o} Iterative Feedback Loop 
& 0.982 & 0.955 & 0.800 & 0.874 & 0.778 & 0.676 & 0.844 & 0.866 & 0.957 & 0.286 & 0.328 & 0.176 & 0.523 \\
\midrule
\multicolumn{14}{l}{\emph{\textbf{Ablation on the RL Data Generation Agent}}} \\
Reduced Diagnostic Criteria & 1.000 & 0.886 & 0.750 & 0.878 & 0.668 & 0.618 & 0.800 & 0.871 & 0.939 & 0.281 & 0.330 & 0.180 & 0.520 \\
Hard-Only Sampling & 1.000 & 0.965 & 0.869 & 0.825 & 0.818 & 0.643 & 0.853 & 0.869 & 0.960 & 0.292 & 0.339 & 0.190 & \textbf{0.530} \\
\bottomrule
\end{tabular}%
}
\end{table*}

\begin{table*}[h]
\centering
\caption{Detailed ablation of the Iterative Feedback Loop and the RL Data Generation Agent on \textbf{DPG} and \textbf{CVTG-2K}.}
\vspace{1mm}
\label{tab:ablation_dpg_cvtg}
\scriptsize
\setlength{\tabcolsep}{3.0pt}
\renewcommand{\arraystretch}{1.12}
\resizebox{\textwidth}{!}{%
\begin{tabular}{l|cccccc|ccccc}
\toprule
\multirow{2}{*}[-1pt]{\textbf{Setting}}
& \multicolumn{6}{c|}{\textbf{DPG}}
& \multicolumn{5}{c}{\textbf{CVTG-2K}} \\
\cmidrule(lr){2-7} \cmidrule(lr){8-12}
& Global & Entity & Attribute & Relation & Other & \textbf{Overall}
& \textbf{Word Accuracy} & NED & CLIPScore & VQAScore & Aesthetic \\
\midrule
APT
& 93.46 & 92.94 & 92.73 & 92.46 & 91.73 & \textbf{87.88}
& \textbf{0.795} & 0.905 & 0.784 & 0.963 & 5.999 \\
\midrule
\multicolumn{12}{l}{\emph{\textbf{Ablation on the Iterative Feedback Loop}}} \\
\textit{w/o} Iterative Feedback Loop
& 94.46 & 91.75 & 92.38 & 93.49 & 88.65 & 87.61
& 0.791 & 0.912 & 0.807 & 0.962 & 5.494 \\
\midrule
\multicolumn{12}{l}{\emph{\textbf{Ablation on the RL Data Generation Agent}}} \\
Reduced Diagnostic Criteria
& 93.77 & 91.74 & 92.13 & 89.53 & 94.54 & 87.53
& 0.763 & 0.900 & 0.796 & 0.963 & 5.769 \\
Hard-Only Sampling
& 93.71 & 93.55 & 91.65 & 91.87 & 89.08 & 87.58
& 0.794 & 0.917 & 0.806 & 0.962 & 5.734 \\
\bottomrule
\end{tabular}%
}
\end{table*}


\clearpage
\section{Visualization}

\begin{figure}[!h]
    \centering
    \includegraphics[width=0.99\textwidth]{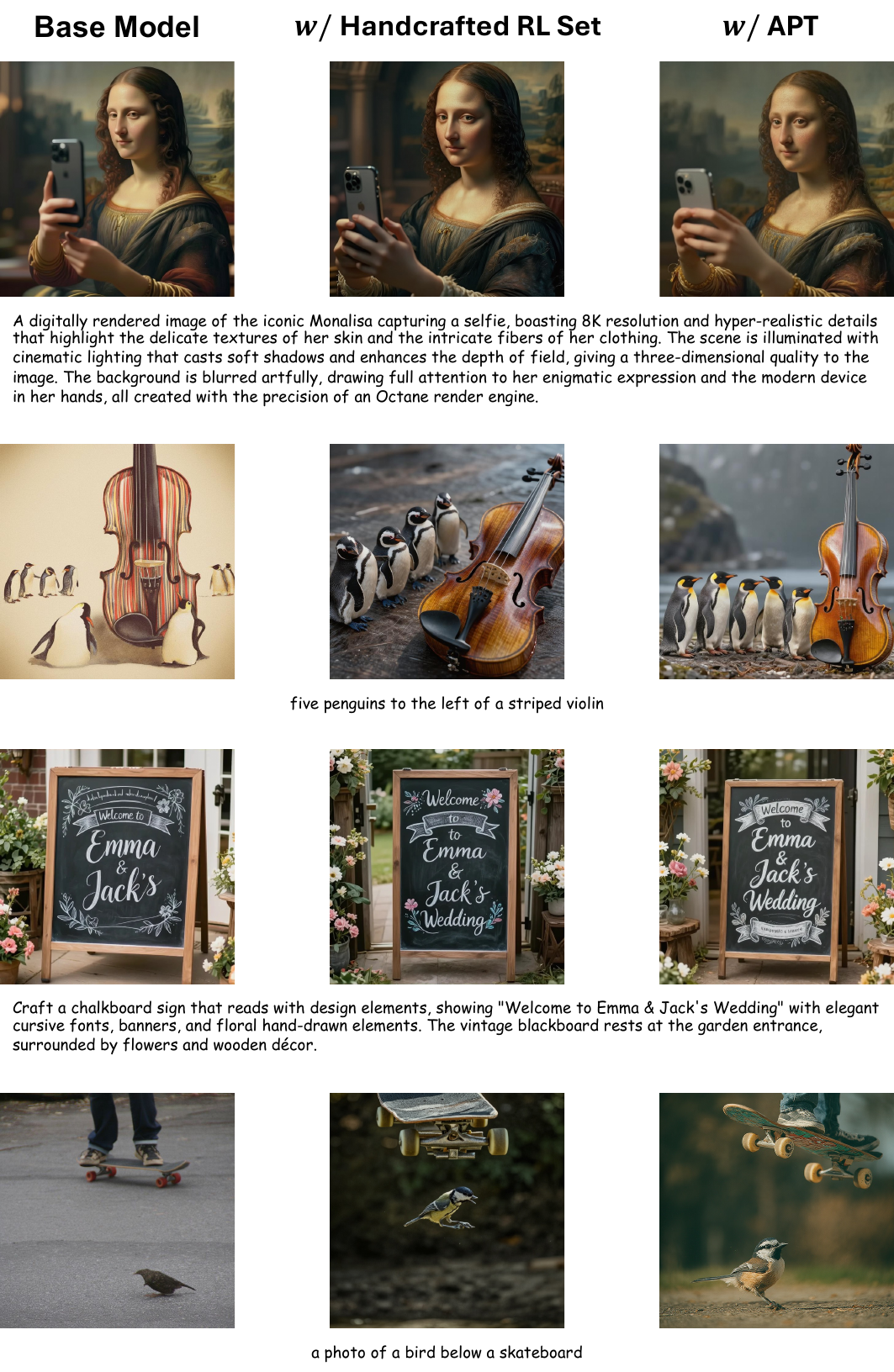}
    \vspace{-3mm}
    \caption{Qualitative comparison of the base model, handcrafted RL baseline, and APT on \textit{Lens}.}
    \vspace{-2mm}
    \label{fig:visual_lens_1}
\end{figure}

\begin{figure}[!h]
    \centering
    \includegraphics[width=0.99\textwidth]{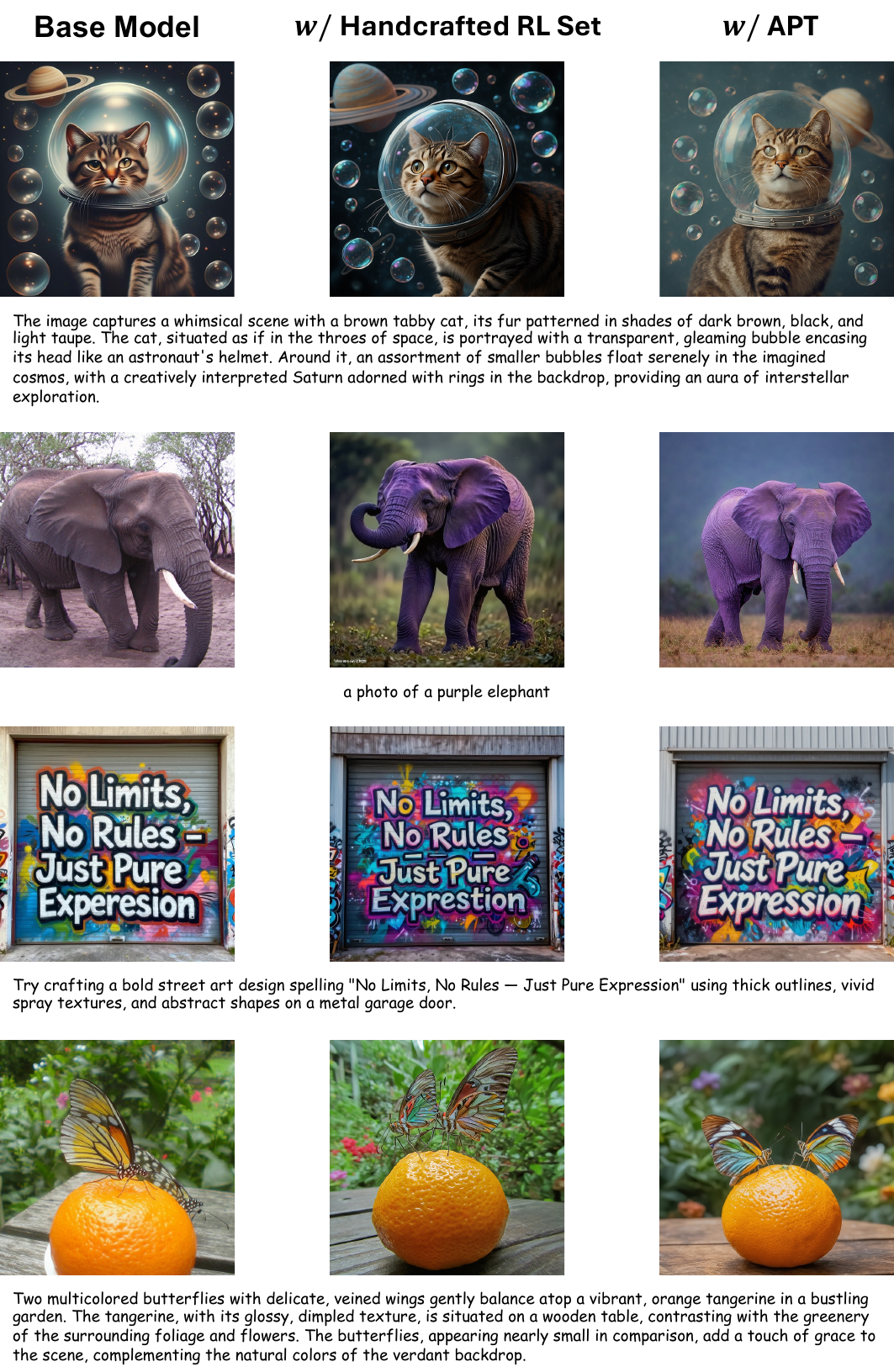}
    \vspace{-3mm}
    \caption{Additional qualitative comparison of the base model, handcrafted RL baseline, and APT on \textit{Lens}.}
    \vspace{-2mm}
    \label{fig:visual_lens_2}
\end{figure}

\begin{figure}[!h]
    \centering
    \includegraphics[width=0.99\textwidth]{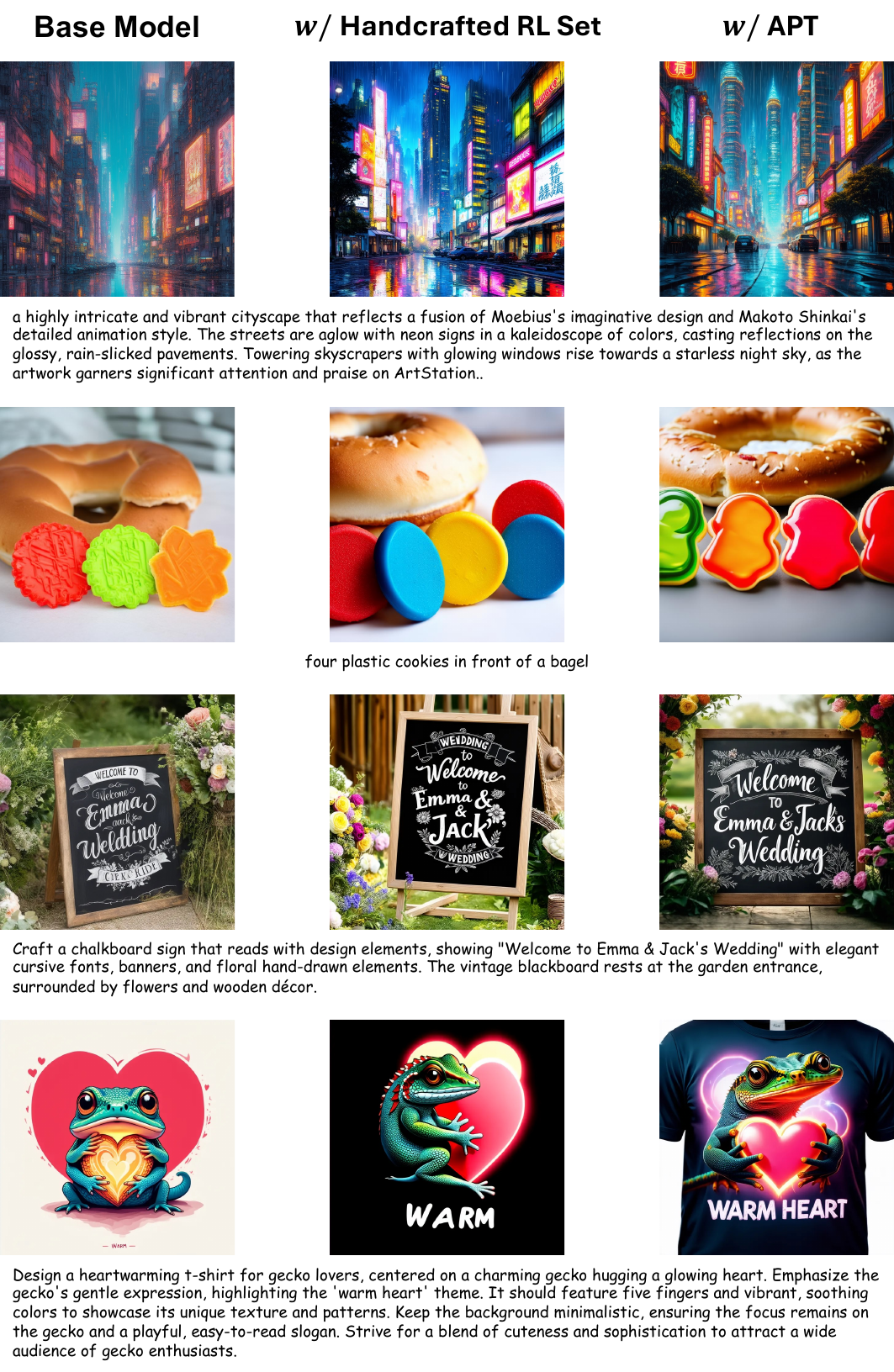}
    \vspace{-3mm}
    \caption{Qualitative comparison of the base model, handcrafted RL baseline, and APT on \textit{SD3.5-Medium}.}
    \vspace{-2mm}
    \label{fig:visual_sd3}
\end{figure}

\begin{figure}[!h]
    \centering
    \includegraphics[width=0.99\textwidth]{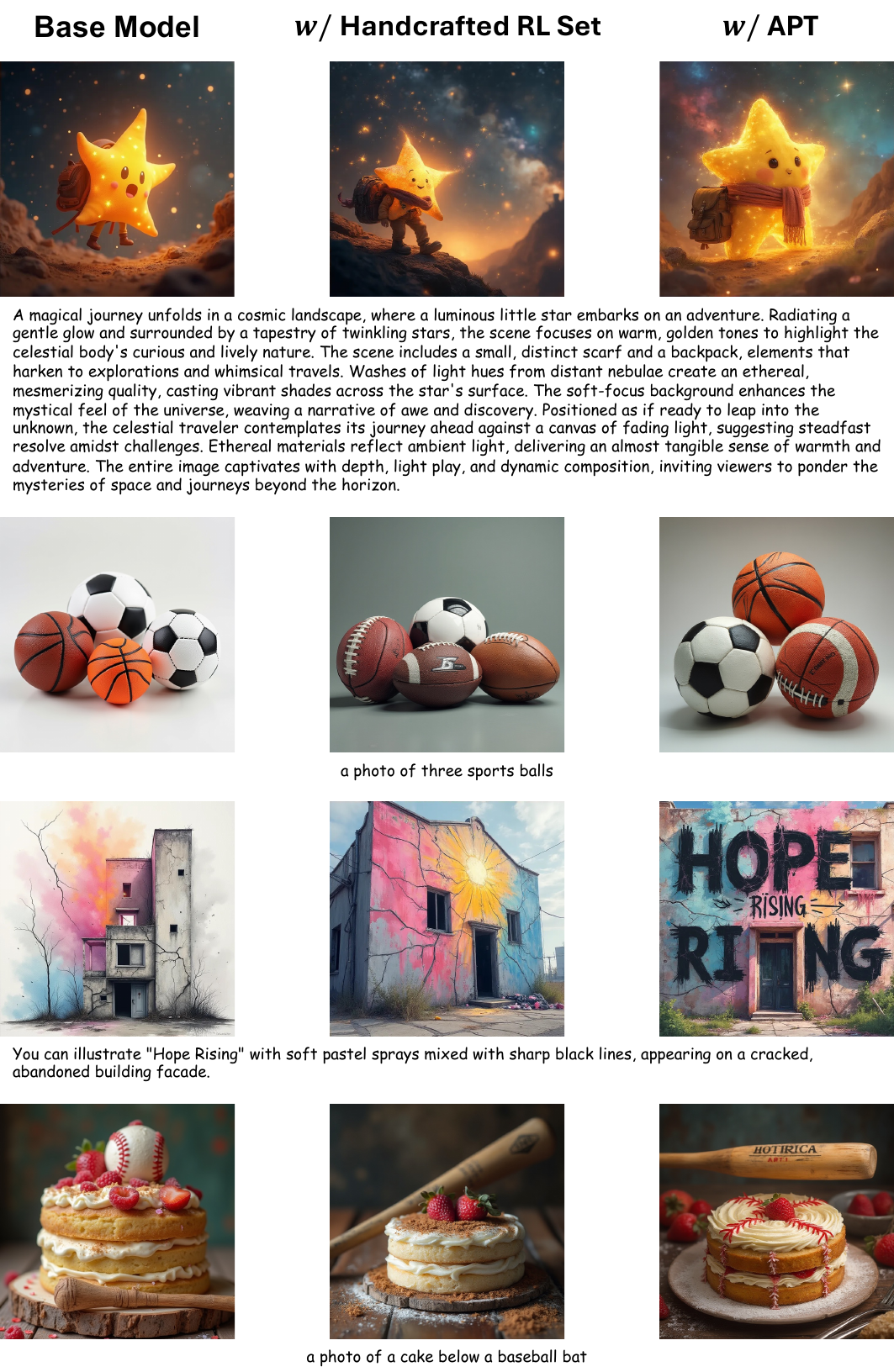}
    \vspace{-3mm}
    \caption{Qualitative comparison of the base model, handcrafted RL baseline, and APT on \textit{FLUX.1[Dev]}.}
    \vspace{-2mm}
    \label{fig:visual_flux}
\end{figure}

\begin{figure}[!h]
    \centering
    \includegraphics[width=0.99\textwidth]{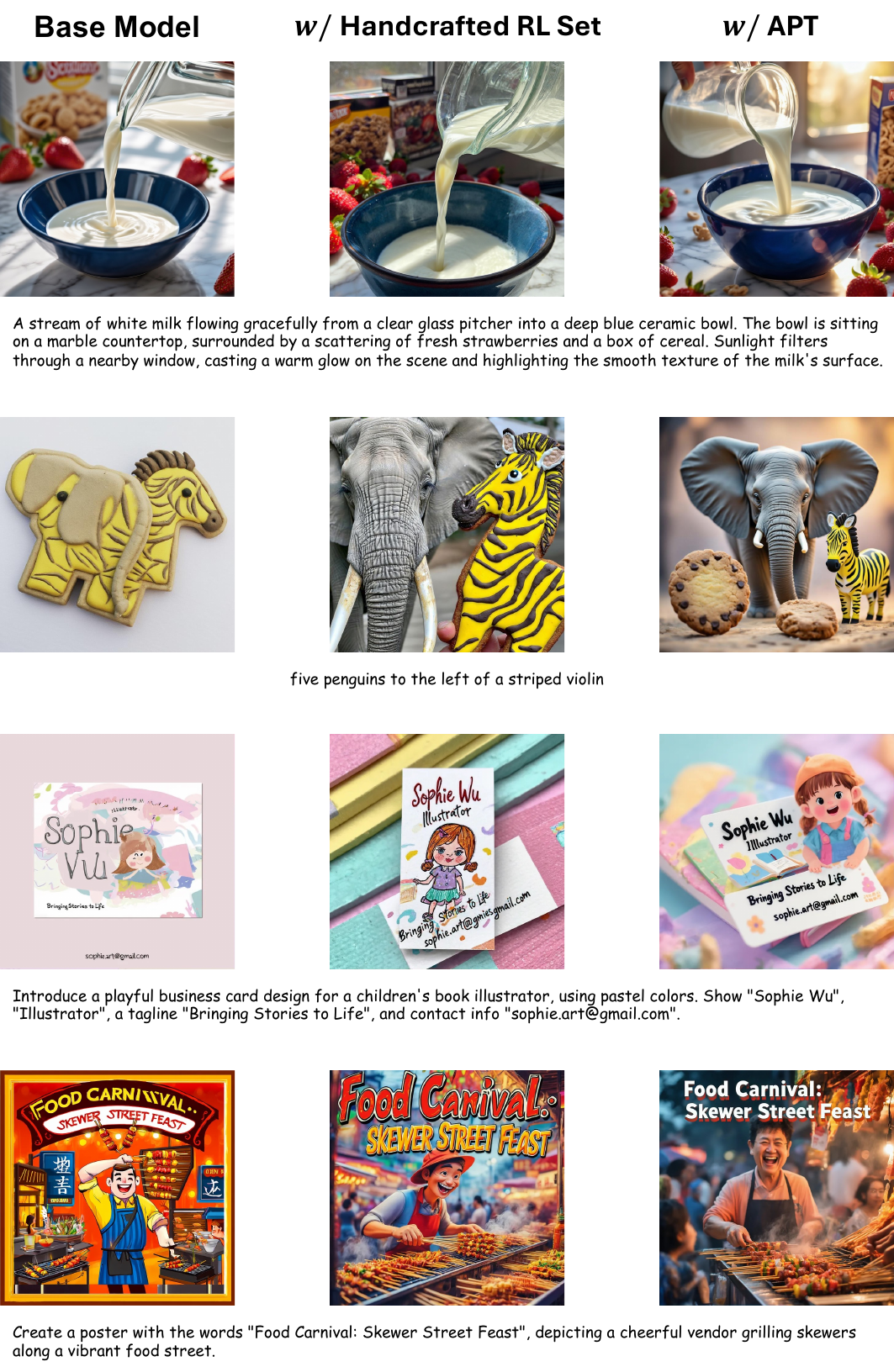}
    \vspace{-3mm}
    \caption{Qualitative comparison of the base model, handcrafted RL baseline, and APT on \textit{LongCat-Image}.}
    \vspace{-2mm}
    \label{fig:visual_longcat}
\end{figure}

\clearpage
\section{Prompt}

\subsection{Prompt for Evaluation Prompt Generation}
\label{sec:prompt-eval-prompt}

\begin{tcolorbox}[
  breakable,
  enhanced,
  colback=white,
  colframe=black!70,
  boxrule=0.8pt,
  arc=2pt,
  title={Prompt for Evaluation Prompt Generation},
  colbacktitle=black!75,
  coltitle=white,
  fonttitle=\bfseries,
  left=8pt,
  right=8pt,
  top=8pt,
  bottom=8pt
]
\footnotesize
\raggedright

You are an expert prompt engineer for photo-realistic image generation.

\medskip

Given an entity (a concept, object, person type, scene, etc.) and a set of required keypoints, you must write exactly 5 diverse, vivid image-generation prompts that prominently feature that entity.

\medskip

\textbf{Keypoint definitions:} When a keypoint is required, you MUST naturally incorporate relevant detail into every prompt:

\begin{itemize}
  \setlength{\itemsep}{2pt}
  \setlength{\topsep}{3pt}
  \setlength{\parsep}{0pt}
  \setlength{\parskip}{0pt}

  \item \textbf{global:} describe the overall scene composition, atmosphere, and environment surrounding the entity.

  \item \textbf{counting:} include specific quantities of objects or subjects (e.g., ``three apples'', ``a pair of shoes'').

  \item \textbf{position:} specify spatial placement or pose of the entity (e.g., ``in the center'', ``on the left side'', ``foreground vs background'').

  \item \textbf{attribute:} describe distinctive physical attributes (e.g., shape, size, material, texture, condition).

  \item \textbf{colors:} mention explicit color details for the entity or surrounding elements (e.g., ``deep crimson'', ``pale blue'').

  \item \textbf{relation:} describe spatial or semantic relationships between the entity and other objects or people in the scene.

  \item \textbf{style:} specify a particular visual or photographic style (e.g., ``cinematic'', ``macro photography'', ``vintage film grain'').

  \item \textbf{text:} include visible text, labels, signs, or typography within the scene (e.g., a storefront sign, a book title, a label on a bottle).
\end{itemize}

\textbf{Rules:}

\begin{enumerate}
  \setlength{\itemsep}{2pt}
  \setlength{\topsep}{3pt}
  \setlength{\parsep}{0pt}
  \setlength{\parskip}{0pt}

  \item Each prompt should be 30--60 words, written as a single descriptive paragraph (no line breaks).

  \item Prompts must be concrete and cinematic: specify subject details, setting, lighting, camera angle, mood, and realistic texture cues.

  \item The 5 prompts should vary in setting, lighting, composition, and mood; avoid repetition.

  \item The entity must be the clear focal subject of every prompt.

  \item ALL required keypoints must be naturally woven into EVERY prompt; do not ignore any.

  \item Return ONLY a valid JSON object with exactly these keys: ``prompt-1'', ``prompt-2'', ``prompt-3'', ``prompt-4'', and ``prompt-5''. No extra keys, no markdown fences, no explanation.
\end{enumerate}

\textbf{Style and length reference:} ONLY use the following examples for tone, structure, and approximate length. Do NOT copy or imitate their specific content, subjects, or settings. Your prompts must be entirely about the given entity.

\begin{itemize}
  \setlength{\itemsep}{4pt}
  \setlength{\topsep}{3pt}
  \setlength{\parsep}{0pt}
  \setlength{\parskip}{0pt}

  \item ``A close-up portrait of a young woman with clear skin, dark almond-shaped eyes, and long black hair, facing the camera directly, neutral expression, soft daylight from a nearby window, natural skin texture, realistic eyelashes, subtle lip color, shallow depth of field, photo-realistic.''

  \item ``A middle-aged man with short salt-and-pepper hair and a trimmed beard, sitting in a leather chair in a quiet study, wearing a navy sweater, warm lamp light illuminating one side of his face, detailed wrinkles around the eyes, thoughtful expression, realistic portrait photography.''

  \item ``A smiling teenage boy standing outdoors on a school campus, wearing a blue hoodie and backpack, slightly messy hair, bright morning light, candid expression, natural skin detail, realistic ears, hands, and facial proportions.''

  \item ``An elderly woman with silver hair tied into a bun, wearing thin round glasses and a beige cardigan, seated by a window with soft afternoon sunlight, expressive wrinkles, gentle smile, realistic age detail in face and hands, warm domestic atmosphere.''

  \item ``A young man with curly hair and freckles, standing in front of a plain gray studio background, wearing a white T-shirt, direct eye contact, relaxed posture, even studio lighting, crisp facial detail, realistic pores and skin texture.''
\end{itemize}

\end{tcolorbox}

\subsection{Prompt for Rubric Generation}
\label{sec:prompt-rubric}

We adopt the following system prompt to generate rubrics for each prompt.

\begin{tcolorbox}[
  breakable,
  enhanced,
  colback=white,
  colframe=black!70,
  boxrule=0.8pt,
  arc=2pt,
  title={Prompt for Rubric Generation},
  colbacktitle=black!75,
  coltitle=white,
  fonttitle=\bfseries,
  left=8pt,right=8pt,top=8pt,bottom=8pt
]
\footnotesize
You are an expert in evaluating the quality of images generated by text-to-image models. A user will provide a text prompt. Your task is to reason about the prompt and identify the key rubrics that real users would use to evaluate the generated image. You should consider the rubric categories listed below **only if** they are relevant to the prompt. 

\medskip

1. "OCR Alignment: If the text prompt explicitly or implicitly implies or requires text to appear in the image, evaluate whether any textual content (e.g., labels, signs, clothing text) within the generated image matches the OCR text that is implied or stated in the prompt. If the text prompt says no OCR text/specific OCR text, you should have a corresponding rubric as well. If the given text prompt (e.g., “a person walking on the beach”) does not contain any OCR text related requirement, please do not return OCR-related rubric." \\
2. "Object Count Consistency: Include this if the prompt explicitly or implicitly specifies the number of key objects or subjects. Assess whether the number of objects described in the prompt is accurately reflected in the image. If no specific object count is mentioned in the text prompt, please do not return this count-related rubric." \\
3. "Attribute Accuracy: Include this if the prompt explicitly describes appearance-level attributes (e.g., color, material, texture, patterns). Examine whether each object's visual attributes match the prompt." \\
4. "Action Accuracy: Include this if the prompt explicitly describes actions or dynamic states (e.g., running, holding, playing). Examine whether the actions and interactions match the prompt." \\
5. "Object Placement and Spatial Reasoning: Include this if the prompt implies or specifies spatial relations between objects. Evaluate whether the spatial relationships or positioning of objects in the image correspond to what is expected from the prompt (e.g., “the dog is sitting under the table”)."
\medskip

Please output up to 10 rubrics in a single JSON object.

\medskip

Each key should be a \textbf{short rubric name} (e.g., "Object Count Consistency"), and each value should be a \textbf{brief explanation}.

\medskip

Valid example:\\
\{"Object Count Consistency": "Verify that exactly one cat is shown.", "OCR Alignment": "Ensure visible text matches prompt description."\}

\medskip

Invalid example (CSV format):\\
"Object Count Consistency","Verify that exactly one cat is shown."\\
Invalid examples (DO NOT DO THESE):\\
- \{"Object counting": "..."\} \hspace{1em} \# INVALID renamed base key\\
- \{"Count Object": "..."\} \hspace{1em} \# INVALID reordered/changed\\
- \{"Object  Count  Consistency": "..."\} \# INVALID spacing changed

\medskip

Strictly follow this format:\\
- A single JSON object (not a list)\\
- Keys and values wrapped in double quotes\\
- do not change the key name\\
- No trailing commas or comments\\
- No line breaks within keys or values

\medskip

Your output must be valid JSON parsable with Python's "json.loads()".
\end{tcolorbox}

\subsection{Prompt for RL Reward Generation}
\label{sec:prompt-reward}
Given a prompt $p$ and its corresponding rubrics, we first generate an image $x$ using the current policy model. We then use the same fixed evaluator $V$ (GPT-4.1-mini) as in Equation~\ref{eq:rubric-feedback} to score each rubric with the following system prompt. These binary scores are aggregated into the scalar training reward $r(x,p)$ used in Equation~\ref{eq:round-objective}.

\begin{tcolorbox}[
  breakable,
  enhanced,
  colback=white,
  colframe=black!70,
  boxrule=0.8pt,
  arc=2pt,
  title={Prompt for RL Reward Generation},
  colbacktitle=black!75,
  coltitle=white,
  fonttitle=\bfseries,
  left=8pt,right=8pt,top=8pt,bottom=8pt
]
\footnotesize
\textbf{System\_prompt}: \\
\indent You are evaluating whether an AI-generated image satisfies the following visual quality criterion.\\
\indent You will receive a user prompt that used to generate the image and a visual quality evaluation criterion.\\
\indent Your job is to return \textbf{1} if the image fully satisfies the criterion, or \textbf{0} if it clearly fails. Do not explain or elaborate.\\
\indent \textbf{Only output: 1 or 0. Do not include explanations, just return the value.}\\
\\
\textbf{Text\_prompt}:\\
\indent The prompt of the generated image that needs to be evaluated is: \{user\_prompt\}\\
\indent Evaluation Criterion (\{key\}):\{criterion\}\\
\indent Scoring Instructions:\\
\indent - Return 1 if the image fully satisfies the criterion.\\
\indent - Return 0 if the image clearly fails.\\
\indent Output: 1 or 0 only.
\end{tcolorbox}

\subsection{Prompt for the RL Data Generation Agent}
\label{sec:prompt-agent}

\begin{tcolorbox}[
  breakable,
  enhanced,
  colback=white,
  colframe=black!70,
  boxrule=0.8pt,
  arc=2pt,
  title={Prompt for the RL Data Generation Agent},
  colbacktitle=black!75,
  coltitle=white,
  fonttitle=\bfseries,
  left=8pt,right=8pt,top=8pt,bottom=8pt
]
\footnotesize
\setlength{\parindent}{0pt}
\setlength{\parskip}{3pt}
\setlength{\baselineskip}{1.08\baselineskip}
\raggedright
Below is a set of evaluation prompts. Each was generated directly into an image by that text-to-image model and scored by a rubric (higher reward = the model handles it well; lower reward = the model handles it poorly):\\
\{testing\_report\}

\medskip

Your task is to analyze the prompts across the full reward spectrum, with primary emphasis on the LOW-reward prompts. Identify the shared surface-level and visual-task characteristics of the kinds of prompts the model struggles with. Do NOT focus primarily on semantic topic similarity. Instead, look for recurring patterns such as:\\
- specific object types the model often fails on \\
- difficult attributes (e.g. color, count, size, shape, texture, material, age, pose, expression) \\
- difficult relations (e.g. left/right, above/below, in front of/behind, overlapping, holding, wearing) \\
- text-rendering demands (e.g. OCR text, signage, labels, logos, clothing text) \\
- visual style or medium (e.g. photorealistic, watercolor, vector art, 3D render, sketch, low-poly, cinematic, flat design)

\medskip

Use both the average reward and the per-criterion failures to infer exactly what kinds of visual generation demands the model handles poorly.

Then generate \{N\_train\} new training prompts with a mixed difficulty distribution: \\
- A clear majority should target the model's weak or recurring failure modes. \\
- A smaller portion should target medium-strength capabilities that still show some instability. \\
- A limited minority should target capabilities the model already handles relatively well, to preserve breadth and avoid overfitting only to the weakest areas. 

Important:\\
- Do not generate prompts only from the single weakest category. \\
- Keep the distribution weighted toward weak capabilities, but include some prompts from stronger areas as well. \\
- The prompts should collectively improve the model in a balanced way while still prioritizing its main weaknesses. 

Requirements for the prompts: \\
- They should target currently weak capabilities, especially recurring failure modes. \\
- Do not merely copy existing prompts; create new prompts that are diverse. \\
- Keep prompts natural and realistic for text-to-image model training. \\
- Make the prompts visually specific and testable. 

Output format: \\
- Output exactly \{N\_train\} lines. \\
- Each line must be valid JSONL: one JSON object per line. \\
- Do not output markdown fences. \\
- Do not output explanations, analysis, headings, bullets, or any text before or after the JSONL. \\
- Each JSON object must contain: \\
-- "id": integer from 1 to \{N\_train\} \\
-- "prompt": the training prompt string 

Example format:\\
\{"id":1,"prompt":"..."\} \\
\{"id":2,"prompt":"..."\}

Return only the JSONL prompt objects.

\end{tcolorbox}

\subsection{Prompt for the RL Data Generation Agent in the Reduced Diagnostic Criteria Ablation}
\label{sec:prompt-agent-ablation-1}

\begin{tcolorbox}[
  breakable,
  enhanced,
  colback=white,
  colframe=black!70,
  boxrule=0.8pt,
  arc=2pt,
  title={Prompt for the RL Data Generation Agent in the \textit{Reduced Diagnostic Criteria} Ablation},
  colbacktitle=black!75,
  coltitle=white,
  fonttitle=\bfseries,
  left=8pt,right=8pt,top=8pt,bottom=8pt
]
\footnotesize
\setlength{\parindent}{0pt}
\setlength{\parskip}{3pt}
\setlength{\baselineskip}{1.08\baselineskip}
\raggedright
Below is a set of evaluation prompts. Each was generated directly into an image by that text-to-image model and scored by a rubric (higher reward = the model handles it well; lower reward = the model handles it poorly):\\
\{testing\_report\}

\medskip

Your task is to analyze the prompts across the full reward spectrum, with primary emphasis on the LOW-reward prompts. Identify the shared surface-level and visual-task characteristics of the kinds of prompts the model struggles with. Do NOT focus primarily on semantic topic similarity. Instead, look for recurring patterns such as:\\
- specific object types the model often fails on \\
- visual style or medium (e.g. photorealistic, watercolor, vector art, 3D render, sketch, low-poly, cinematic, flat design) 

\medskip

Use both the average reward and the per-criterion failures to infer exactly what kinds of visual generation demands the model handles poorly.

Then generate \{N\_train\} new training prompts with a mixed difficulty distribution: \\
- A clear majority should target the model's weak or recurring failure modes. \\
- A smaller portion should target medium-strength capabilities that still show some instability. \\
- A limited minority should target capabilities the model already handles relatively well, to preserve breadth and avoid overfitting only to the weakest areas. 

Important:\\
- Do not generate prompts only from the single weakest category. \\
- Keep the distribution weighted toward weak capabilities, but include some prompts from stronger areas as well. \\
- The prompts should collectively improve the model in a balanced way while still prioritizing its main weaknesses. 

Requirements for the prompts: \\
- They should target currently weak capabilities, especially recurring failure modes. \\
- Do not merely copy existing prompts; create new prompts that are diverse. \\
- Keep prompts natural and realistic for text-to-image model training. \\
- Make the prompts visually specific and testable. 

Output format: \\
- Output exactly \{N\_train\} lines. \\
- Each line must be valid JSONL: one JSON object per line. \\
- Do not output markdown fences. \\
- Do not output explanations, analysis, headings, bullets, or any text before or after the JSONL. \\
- Each JSON object must contain: \\
-- "id": integer from 1 to \{N\_train\} \\
-- "prompt": the training prompt string 

Example format:\\
\{"id":1,"prompt":"..."\} \\
\{"id":2,"prompt":"..."\}

Return only the JSONL prompt objects.

\end{tcolorbox}

\subsection{Prompt for the RL Data Generation Agent in the Hard-Only Sampling Ablation}
\label{sec:prompt-agent-ablation-2}

\begin{tcolorbox}[
  breakable,
  enhanced,
  colback=white,
  colframe=black!70,
  boxrule=0.8pt,
  arc=2pt,
  title={Prompt for the RL Data Generation Agent in the \textit{Hard-Only Sampling} Ablation},
  colbacktitle=black!75,
  coltitle=white,
  fonttitle=\bfseries,
  left=8pt,right=8pt,top=8pt,bottom=8pt
]
\footnotesize
\setlength{\parindent}{0pt}
\setlength{\parskip}{3pt}
\setlength{\baselineskip}{1.08\baselineskip}
\raggedright
Below is a set of evaluation prompts. Each was generated directly into an image by that text-to-image model and scored by a rubric (higher reward = the model handles it well; lower reward = the model handles it poorly):\\
\{testing\_report\}

\medskip

Your task is to analyze the prompts across the full reward spectrum, with primary emphasis on the LOW-reward prompts. Identify the shared surface-level and visual-task characteristics of the kinds of prompts the model struggles with. Do NOT focus primarily on semantic topic similarity. Instead, look for recurring patterns such as:\\
- specific object types the model often fails on \\
- difficult attributes (e.g. color, count, size, shape, texture, material, age, pose, expression) \\
- difficult relations (e.g. left/right, above/below, in front of/behind, overlapping, holding, wearing) \\
- text-rendering demands (e.g. OCR text, signage, labels, logos, clothing text) \\
- visual style or medium (e.g. photorealistic, watercolor, vector art, 3D render, sketch, low-poly, cinematic, flat design)

\medskip

Use both the average reward and the per-criterion failures to infer exactly what kinds of visual generation demands the model handles poorly.

Then generate \{N\_train\} new training prompts that exclusively target the model's weak or recurring failure modes. Prioritize the most frequent, severe, and persistent errors observed in the evaluation results, and avoid generating prompts that primarily assess capabilities the model already performs well on.

Requirements for the prompts: \\
- They should target currently weak capabilities, especially recurring failure modes. \\
- Do not merely copy existing prompts; create new prompts that are diverse. \\
- Keep prompts natural and realistic for text-to-image model training. \\
- Make the prompts visually specific and testable.

Output format: \\
- Output exactly \{N\_train\} lines. \\
- Each line must be valid JSONL: one JSON object per line. \\
- Do not output markdown fences. \\
- Do not output explanations, analysis, headings, bullets, or any text before or after the JSONL. \\
- Each JSON object must contain: \\
-- "id": integer from 1 to \{N\_train\} \\
-- "prompt": the training prompt string 

Example format:\\
\{"id":1,"prompt":"..."\} \\
\{"id":2,"prompt":"..."\}

Return only the JSONL prompt objects.

\end{tcolorbox}